\documentclass[12pt,a4paper]{article}

\usepackage[T1]{fontenc}
\usepackage{lmodern}
\usepackage[hmargin=28mm,vmargin=32mm]{geometry}
\usepackage[onehalfspacing]{setspace}
\usepackage{amsmath,amssymb,amsthm}
\usepackage{graphicx}
\usepackage{array}
\usepackage[round,authoryear]{natbib}
\usepackage{authblk}
\usepackage{microtype}
\usepackage{etoolbox}
\usepackage{xurl}
\usepackage[hidelinks]{hyperref}

\date{}
\theoremstyle{plain}
\newtheorem{theorem}{Theorem}
\newtheorem{proposition}[theorem]{Proposition}
\theoremstyle{remark}

\newtheorem{remark}{Remark}
\theoremstyle{definition}

\usepackage{algorithm}
\usepackage{algpseudocode}
\usepackage{booktabs}
\usepackage{placeins}
\AtEndEnvironment{figure}{\vspace{-3pt}}
\AtEndEnvironment{table}{\vspace{-3pt}}
\AtBeginDocument{
	\setlength{\abovedisplayskip}{7pt plus2pt minus3pt}
	\setlength{\belowdisplayskip}{7pt plus2pt minus3pt}
	\setlength{\abovedisplayshortskip}{3pt plus2pt}
	\setlength{\belowdisplayshortskip}{5pt plus2pt minus2pt}
}
\allowdisplaybreaks[4]
\newtheorem{assumption}{}

\newcommand{\dl}{\mathrm{d}}
\newcommand{\E}{\mathbb{E}}

\begin{document}

	\raggedbottom

	\title{Metric Conformal Prediction via the\\Conditional Expected Local Radius}

	\author[1]{Rui Qiu}
	\author[2]{Fang Yao}
	\author[1]{Zhou Yu}
	\affil[1]{School of Statistics, KLATASDS-MOE, East China Normal University}
	\affil[2]{School of Mathematical Sciences, Center for Statistical Science, Peking University}

	\maketitle
	\begingroup
	\renewcommand{\thefootnote}{\fnsymbol{footnote}}
	\endgroup

	\begin{abstract}
We introduce the conditional expected local radius (CELR) and develop a new conformal prediction method for non-Euclidean responses in general metric spaces. CELR measures local conditional probability concentration, allowing prediction sets to accommodate heterogeneity and multimodality without requiring densities or a vector space structure. Its conditional distribution yields a nonconformity score that is comparable across covariate values and avoids costly transport comparisons between distance profiles. We further adapt a metric distributional random forest that handles multivariate covariates and nonlinear interactions, while a single forest fit supplies all score components, reducing computation by eliminating repeated local fitting. We establish uniform consistency of the estimators and asymptotic conditional validity under suitable regularity conditions, together with finite-sample marginal coverage under exchangeability. Numerical studies demonstrate effective uncertainty quantification across different metric spaces and substantial computational savings over the conditional profile score method.
	\end{abstract}

	\noindent\textbf{Keywords:} Conditional validity, Conformal prediction, Metric space, Random forest, Random object

	\section{Introduction}\label{sec:introduction}
	Conformal prediction (CP) quantifies predictive uncertainty through prediction sets with finite-sample and distribution-free coverage guarantees \citep{vovk2005algorithmic}. In regression, the aim is to construct a set $\hat{C}_{1-\alpha}(X_0)$ for a future observation $(X_0,Y_0)$ satisfying the marginal coverage requirement
	\begin{equation}\label{marginal validity}
		\mathbb{P}\bigl( Y_0 \in \hat{C}_{1-\alpha}(X_0) \bigr) \ge 1-\alpha.
	\end{equation}
	Under exchangeability, this guarantee holds without specifying a parametric model for the joint distribution of $(X,Y)$ \citep{shafer2008tutorial}. For scalar responses, CP yields prediction intervals and bands \citep{lei2014distribution, lei2018distribution, chernozhukov2021distributional, sesia2021conformal, barber2023conformal}. The nonconformity score ranks candidate responses and determines the size and shape of the prediction set. Studentized residual scores \citep{lei2018distribution} and scores based on conditional quantiles \citep{romano2019conformalized}, for example, allow prediction intervals to adapt to heteroskedasticity. Other developments concern data reuse \citep{vovk2015cross, barber2021predictive}, adaptation to changing distributions \citep{gibbs2021adaptive, gibbs2024conformal}, and the selection and aggregation of prediction sets \citep{yang2025selection,xu2026aggregating}.
	
	Marginal coverage averages over the covariate distribution and need not provide adequate coverage at a particular covariate value. Exact conditional validity without restrictions on the joint distribution is generally unattainable \citep{vovk2012conditional, lei2014distribution, barber2021predictive}. A common alternative is asymptotic conditional validity:
	\begin{equation}\label{conditional validity}
		\mathbb{P}\bigl( Y_0 \in \hat{C}_{1-\alpha}(X_0) \mid X_0 \bigr) = 1-\alpha + o_p(1).
	\end{equation}
	To attain \eqref{conditional validity}, \citet{chernozhukov2021distributional} used conditional distribution ranks in place of residual scores \citep{lei2018distribution}, obtaining prediction intervals. Conditional density ranks yield sets that converge to highest predictive density regions and adapt to skewness, heavy tails, and multimodality \citep{izbicki2022cd}.
	Localized conformal methods pursue related coverage goals \citep{guan2023localized, hore2025conformal, gibbs2025conformal}. For a general account of conformal prediction, see \citet{angelopoulos2024theoretical}.
	
	These methods are designed primarily for Euclidean responses. Non-Euclidean responses, such as phylogenetic trees, covariance matrices, and networks, are naturally treated as random objects in metric spaces.
	Methods for such data include Fr\'echet regression and regression on manifolds \citep{lin2017extrinsic, petersen2019frechet, dubey2020functional, lin2021total, qiu2024random, bhattacharjee2025nonlinear}, dimension reduction \citep{panaretos2014principal, virta2022sliced, zhang2024dimension}, and deep learning \citep{iao2025deep, kim2025dfnn, zhou2026end, matabuena2026deep}. Constructing informative prediction sets requires additional information about the spread and shape of the conditional law. Vector operations and densities need not be available in these spaces, so conventional constructions based on residuals or densities are not directly applicable. The conditional law may also concentrate in several separated regions whose locations and relative masses vary with the predictors. Such multimodality motivates prediction sets that adapt to the geometry of the conditional distribution.
	
	\citet{lugosi2025conformal} developed conformal and locally adaptive $k$-nearest-neighbor methods for uncertainty quantification in metric spaces. Their framework accommodates diverse regression estimators and provides interpretable prediction regions for non-Euclidean responses. \citet{zhou2025conformal} constructed conformal prediction sets from conditional distance profiles, using the average transport cost between profiles. This profile-based construction can handle multimodal conditional distributions without requiring a response density. A candidate's distance profile can also be summarized by the radii needed to accumulate probability mass around it. These local radii underlie the distance-to-a-measure (DTM) family \citep{chazal2011geometric, chazal2016rates}, including the $L^1$ formulation studied by \citet{brecheteau2019statistical},  which primarily concern unconditional geometric or topological inference and comparisons between metric spaces equipped with probability measures.

	In this work, our contributions are three-fold. 
	First, we construct a conformal score from the conditional expected local radius (CELR) that is distinct from the conditional distance profiles. Its unconditional counterpart, the expected local radius (ELR), coincides with the $L^1$ formulation of DTM studied by \citet{brecheteau2019statistical}, up to the convention used at probability mass thresholds. Under the conditional law of $Y\mid X=x$, CELR averages the radii needed to accumulate probability mass around a candidate. This description of local concentration complements metric depths, ranks, and quantiles \citep{geenens2023statistical, dai2023tukey, liu2022quantiles, dubey2024metric}. We transform CELR by its conditional distribution before conformal calibration, making the score comparable across covariate values. The resulting prediction sets can accommodate heterogeneity and multimodality in the conditional law without requiring densities or a vector space structure. CELR also simplifies computation: it uses only the candidate's conditional distance profile and avoids the transport comparisons between profiles required by \citet{zhou2025conformal}.
	
	Second, we estimate the conditional law with a metric distributional random forest (MDRF). Local linear smoothing often requires additional structure or dimension reduction as the number of covariates increases \citep{fan1996local}. Our construction uses maximum mean discrepancy to select splits and draws on generalized random forests \citep{athey2019generalized, wager2018estimation} and distributional random forests \citep{cevid2022distributional}. It differs from the local linear approach of \citet{zhou2025conformal} in three respects:
	
	\begin{enumerate}
		\item[(1)] \emph{Unified estimation.} 
		A single forest fit provides the weights for estimating conditional distance profiles, CELRs, and the conditional distributions of CELRs. Reusing these weights reduces the computation needed at each stage. \citet{zhou2025conformal} estimate distance profiles, average transport costs, and their conditional distributions through separate local linear fits.
		
		\item[(2)] \emph{Adaptive weights.} 
		The forest uses response information to choose splits, so its weights can adapt to variation in the conditional law \citep{lin2006random, qiu2024random}. The local linear weights in \citet{zhou2025conformal} are instead based on Euclidean distances in covariate space.
		
		\item[(3)] \emph{Multivariate covariates.}
		The forest partitions the multivariate covariate space directly and can represent nonlinear interactions and local variation. The multivariate procedure of \citet{zhou2025conformal} relies on a single index assumption, which can be restrictive when covariate effects cannot be captured by a single linear combination.
	\end{enumerate}
	
	Third, we establish uniform consistency of the estimated conditional distance profiles, CELRs, and score ranks, and derive asymptotic conditional validity. Pointwise consistency does not suffice because the score construction evaluate fitted quantities at arguments that depend on the data. The proof treats the forest as a growing-order U-statistic whose order increases with sample size. We derive a maximal inequality for the associated U-process and combine it with leaf shrinkage and stability bounds to control the accumulated estimation error.
	
	The paper is organized as follows. Section~\ref{sec:local_concentration} defines ELR and studies its geometric properties. Section~\ref{sec:metric_comformal_prediction} presents the score construction, split conformal calibration, and forest estimator. Section~\ref{sec:theory} gives the uniform consistency and asymptotic conditional validity results. Sections~\ref{sec:simulation_studies} and~\ref{sec:real_data_analysis} present simulations in several metric spaces and two data applications, respectively. Section~\ref{sec:discussion} concludes. Proofs and additional implementation details are in the supplementary material.

	\section{Expected local radius}
	\label{sec:local_concentration}
	Let $(\Omega, d)$ be a metric space, and write $B(\omega,r)=\{y\in\Omega:d(y,\omega)\le r\}$ for the closed ball of radius $r$ centered at $\omega\in\Omega$. The distance profile at $\omega$ \citep{dubey2024metric} is the distribution function of $d(Y,\omega)$:
	$$
	F(\omega, r):=\mathbb{P}\{Y\in B(\omega,r)\}=\mathbb{P}\{d(\omega, Y) \leq r\}, \quad r \geq 0.
	$$
	The corresponding quantile function is
	$$
	Q(\omega,\tau) := \inf \left\{ r \geq 0 : F(\omega, r) \geq \tau \right\}, \quad \tau \in (0,1].
	$$
	When finite, $Q(\omega,\tau)$ is the smallest radius of a ball centered at $\omega$ that contains probability mass of at least $\tau$.
	For a locality level $\beta \in (0, 1]$, define the expected local radius (ELR) at $\omega$ by
	\begin{align*}
		R_{\beta}(\omega) := \frac{1}{\beta} \int_0^\beta Q(\omega,\tau) \, \dl \tau.
	\end{align*}
	ELR measures concentration around $\omega$ by averaging distance quantiles over probability levels up to $\beta$. Smaller values indicate greater concentration at this scale.
	
	ELR belongs to the DTM family. \citet{chazal2011geometric} defined local mass radii using a strict mass inequality and combined them through an $L^2$ average. The $L^1$ version studied by \citet{brecheteau2019statistical} agrees with $R_\beta(\omega)$; the different conventions at mass thresholds do not affect the integral. We use the name expected local radius to emphasize its geometric interpretation.
	Both ELR and expected shortfall are averages of quantiles: ELR uses the lower tail of a distance distribution, and expected shortfall uses the upper tail of a loss distribution. Supplementary Section~S.1 compares ELR with classical statistics based on quantiles.
	
	Using ELR in a conformal score avoids the transport comparisons between conditional distance profiles required by the method of \citet{zhou2025conformal}. The following proposition gives three basic properties. Monotonicity and the quantile bounds follow from quantile averaging; Lipschitz continuity is a standard DTM property \citep{chazal2011geometric, brecheteau2019statistical}.
	
	\begin{proposition}[Basic properties of ELR]
		\label{basic_properties}
		For $\beta=1$, assume that $R_1(\omega)<\infty$ for all $\omega\in\Omega$.
		The ELR satisfies the following properties:
		\begin{enumerate}
			\item \textbf{(Monotonicity).} For every $\omega\in\Omega$, the map $\beta\mapsto R_\beta(\omega)$ is nondecreasing on $(0,1]$.
			\item \textbf{(Quantile sandwich).} For every $\omega\in\Omega$ and $0<\beta_0<\beta\le1$,
			$$
			\frac{\beta-\beta_0}{\beta}\,Q(\omega, \beta_0)\ \le\ R_\beta(\omega)\ \le\ Q(\omega, \beta).
			$$
			\item \textbf{(Lipschitz continuity).} For every $\beta\in(0,1]$ and $\omega,\omega'\in\Omega$,
			$$
			\big|R_\beta(\omega)-R_\beta(\omega')\big|
			\ \le\ d(\omega,\omega').
			$$
		\end{enumerate}
	\end{proposition}
	
	Proposition~\ref{basic_properties} relates ELR to ordinary distance quantiles. Its Lipschitz bound shows that moving the candidate center changes ELR by at most the distance moved.
	
	The role of $\beta$ can be seen from its limiting cases. When $\beta = 1$ and the first distance moment is finite,
	$$
	R_1(\omega) = \int_0^1 Q(\omega, \tau)\, \dl \tau = \mathbb{E}[d(Y, \omega)],
	$$
	so ELR averages distances over the whole distribution. As $\beta\downarrow0$, $R_\beta(\omega)$ approaches the distance from $\omega$ to the support of $Y$. The next proposition gives the rate of decay on a Riemannian manifold when the density near $\omega$ is continuous and positive.
	
	\begin{proposition}[Metric concentration via density]
		\label{prop:inverse_density}
		Let $(\Omega,d)$ be a $k$-dimensional Riemannian manifold equipped with its geodesic distance. Fix an interior point $\omega \in \Omega$ and assume that $P_Y$ admits a continuous and strictly positive density $f$ in a neighborhood of $\omega$ with respect to Riemannian volume. Then, as $\beta \rightarrow 0$,
		$$
		R_\beta(\omega)
		=
		\frac{k}{k+1}
		\left(\frac{1}{V_k f(\omega)}\right)^{1/k}
		\beta^{1/k}
		+
		o\!\left(\beta^{1/k}\right),
		$$
		where $V_k$ is the volume of the unit ball in $\mathbb{R}^k$.
	\end{proposition}
	
	Proposition~\ref{prop:inverse_density} shows that the rescaled ELR $\beta^{-1/k}R_\beta(\omega)$ is asymptotically proportional to $f(\omega)^{-1/k}$. As $\beta\downarrow0$, higher local density corresponds to smaller rescaled ELR values. This motivates using ELR to rank candidate responses by local concentration. In a general metric space, ELR is still defined by metric-ball probabilities, even when no natural density is available. The next proposition gives the corresponding expansion directly in terms of these probabilities.
	
	\begin{proposition}[Metric concentration via small-ball probability]\label{prop:metric_concentration}
		Let $(\Omega,d)$ be a general metric space. Assume that, at a point $\omega\in\Omega$, the distribution of $Y$ satisfies the local small-ball expansion
		$$
		\mathbb{P}\bigl(d(Y,\omega)\le r\bigr)
		=
		a(\omega)\, r^{\gamma(\omega)}\bigl(1+o(1)\bigr),
		\qquad r\to 0,
		$$
		where $\gamma(\omega)>0$ is the local dimension exponent and $a(\omega)>0$ is the local concentration coefficient. Then, as $\beta \rightarrow 0$,
		$$
		R_\beta(\omega)
		=
		\frac{\gamma(\omega)}{\gamma(\omega)+1}
		\left( \frac{1}{a(\omega)} \right)^{1/\gamma(\omega)}
		\beta^{1/\gamma(\omega)}
		+
		o\!\left(\beta^{1/\gamma(\omega)}\right).
		$$
	\end{proposition}
	
\begin{figure}[ht]
		\centering
		\includegraphics[width=0.32\linewidth]{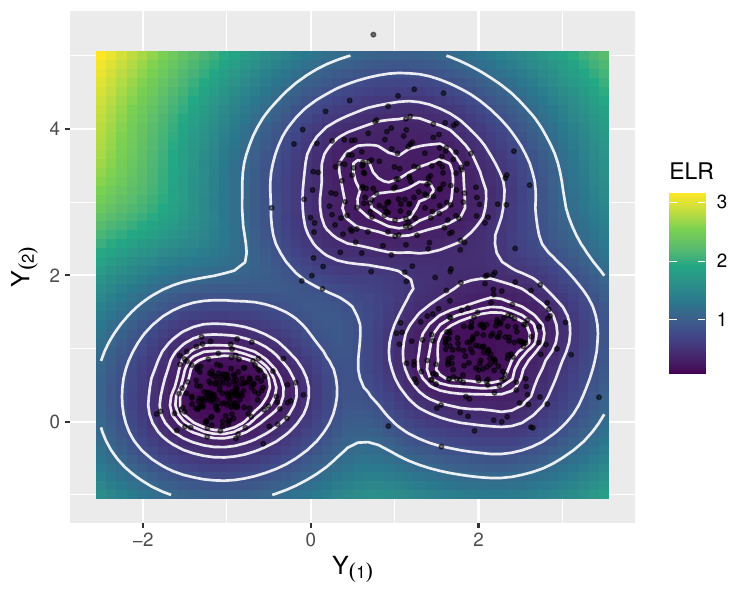}
		\includegraphics[width=0.32\linewidth]{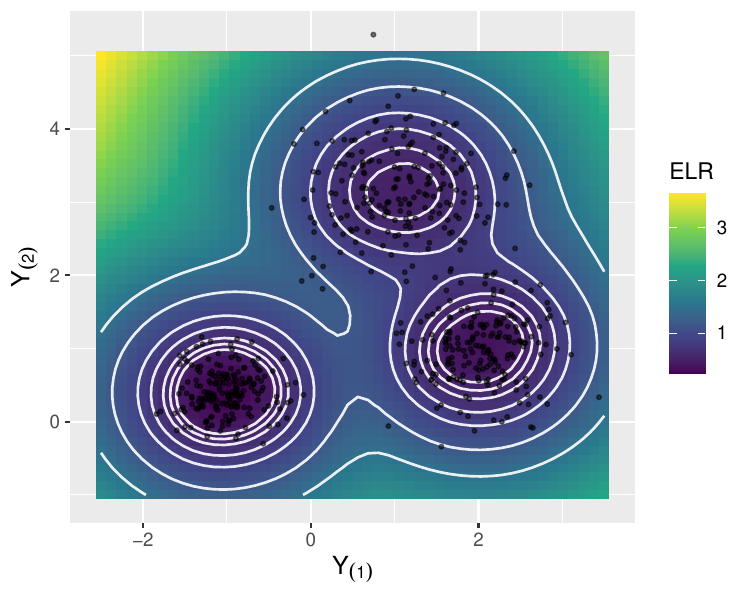}
		\includegraphics[width=0.32\linewidth]{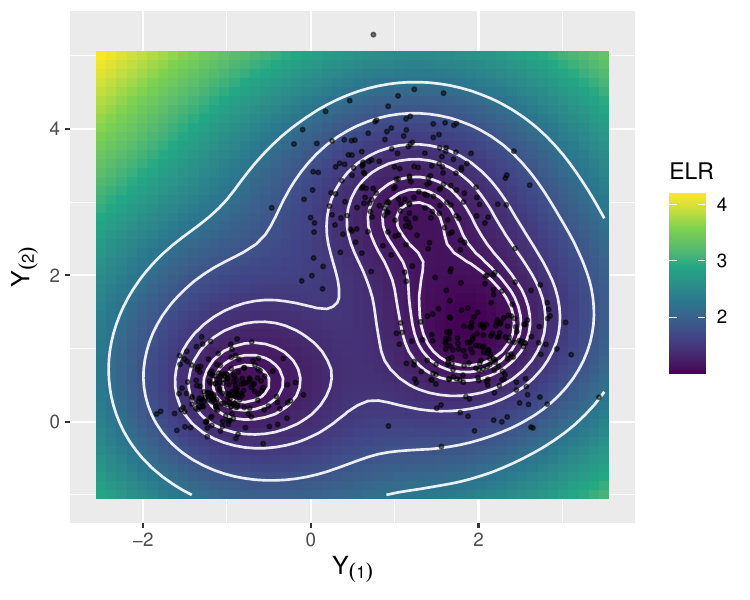}
		\caption{Empirical surfaces of the ELR functional $\omega\mapsto R_\beta(\omega)$ for a three-component Gaussian mixture with $\beta=0.05,0.2$, and $0.5$. Smaller values of $\beta$ emphasize fine local concentration, whereas larger values shift the functional toward global dispersion. Colors represent ELR values, white curves show level sets, and black points denote observations.}
		\label{beta_value}
	\end{figure}

	Smaller values of $\beta$ emphasize probability mass near $\omega$, while larger values include mass farther away. Very small values can give unstable estimates because only a few observations contribute to the empirical average. In the numerical studies, we consider candidate values of $\beta$ in $[0.1,0.25]$ to balance locality and robustness. Figure~\ref{beta_value} illustrates the effect of $\beta$ for a Gaussian mixture with three components.
	
	\section{ELR-based conformal prediction in metric spaces}
	\label{sec:metric_comformal_prediction}
	
	ELR quantifies local concentration around any point in a metric space. We now use its conditional counterpart to construct prediction sets for random objects.
	
	\subsection{Split conformal prediction}
	Let $(X, Y)$ take values in $\mathcal{X} \times \Omega$, where $\mathcal{X}=[0,1]^p$ and $(\Omega, d)$ is a Polish metric space equipped with its Borel $\sigma$-field. For $x \in \mathcal X$ and $\omega \in \Omega$, the conditional distance profile (CDP) is
	$$
	F_{x}(\omega, r) := \mathbb{P}\left\{ d(Y, \omega) \leq r \mid X = x \right\}, \quad r \ge 0.
	$$
	The corresponding quantile function is
	$$
	Q_{x}(\omega, \tau) := \inf \left\{ r \geq 0 : F_{x}(\omega, r) \geq \tau \right\}, \quad \tau \in (0,1].
	$$
	For a locality level $\beta \in (0,1]$, define the conditional expected local radius (CELR) by
	\begin{align*}
		R_{\beta}(\omega \mid x) := \frac{1}{\beta} \int_0^\beta Q_{x}(\omega, \tau) \, \dl \tau.
	\end{align*}
	This is the ELR under the conditional law of $Y$ given $X=x$.
	
	Given an independent and identically distributed (i.i.d.) sample $(X_i, Y_i) \in \mathcal X \times \Omega$, $i =1, \ldots, 2n$, we seek a prediction set $\hat{C}_{1-\alpha}(X_0)$ satisfying \eqref{conditional validity} for an additional i.i.d. observation $(X_0,Y_0)$.
	At the chosen locality level $\beta$, smaller CELR values indicate greater concentration of conditional probability mass around a candidate response, while larger values indicate weaker concentration. This motivates using CELR as a nonconformity measure. Because its scale may vary with $x$, we transform CELR by its conditional distribution function \citep{chernozhukov2021distributional, izbicki2022cd, zhou2025conformal}:
	\begin{align*}
		S(z \mid x):= \mathbb{P}\left\{ R_{\beta}(Y \mid X) \leq z \mid X = x \right\}.
	\end{align*}
	We call $S\{R_\beta(Y_0 \mid X_0)\mid X_0\}$ the conditional radius score (CRS) and use it as the oracle nonconformity score. If the conditional distribution of $R_{\beta}(Y_0 \mid X_0)$ given $X_0$ is continuous, the probability integral transform implies that
	$
	S\left(R_\beta\left(Y_{0} \mid X_{0}\right) \mid X_{0}\right)
	$
	is uniformly distributed on $[0,1]$ and independent of $X_{0}$.
	The oracle prediction set
	$$C^{*}_{1-\alpha}(x) := \{\omega \in \Omega : S(R_{\beta}(\omega \mid x) \mid x) \leq 1-\alpha\}$$
	therefore satisfies
	$\mathbb{P}\{Y_0 \in C^{*}_{1-\alpha}(X_0) \mid X_0\} = 1-\alpha$.
	
	In practice, the conditional law of $Y \mid X=x$ is unknown. We use split conformal prediction \citep{lei2018distribution}, which separates model fitting from calibration. Randomly partition $\{(X_i, Y_i)\}_{i=1}^{2n}$ into a training set $\mathcal{D}_{\mathrm{tra}}=\{(X_i,Y_i):i\in  \mathcal{I}_{\mathrm{tra}}\}$ and a calibration set $\mathcal{D}_{\mathrm{cal}}=\{(X_i,Y_i):i\in  \mathcal{I}_{\mathrm{cal}}\}$, where $\mathcal{I}_{\mathrm{tra}} \cup \mathcal{I}_{\mathrm{cal}} = \{1, \ldots, 2n\}$. Use only $\mathcal{D}_{\mathrm{tra}}$ to construct the estimators $\hat{F}_x(\omega,\cdot)$, $\hat{R}_{\beta}(\omega \mid x)$, and $\hat{S}(z \mid x)$.
	For each $i \in \mathcal{I}_{\mathrm{cal}}$, compute $\hat{S}_i = \hat{S}\{\hat{R}_{\beta}(Y_{i} \mid X_{i}) \mid X_{i}\}$. Let $\hat{Q}_{1-\alpha}$ be the $(1-\alpha)(1 + 1 / |\mathcal{D}_{\mathrm{cal}}|)$ empirical quantile of $\{\hat S_i:i\in\mathcal I_{\mathrm{cal}}\}$. The resulting prediction set at $X_{0}$ is
	$$
	\hat{C}_{1-\alpha}(X_{0}) =
	\Big\{\omega\in\Omega:\ \hat{S}\big(\hat{R}_{\beta}(\omega \mid X_{0}) \mid X_{0}\big)\le \hat{Q}_{1-\alpha}\Big\}.
	$$
	Conditional on the training sample and the data split, the calibration and test scores are exchangeable. This gives the marginal coverage guarantee \eqref{marginal validity}.
	
	\begin{theorem}[Finite-sample marginal coverage]\label{thm:marginal}
		The split conformal set $\hat{C}_{1-\alpha}(X_{0})$ satisfies
		\[
		\mathbb{P}\left\{Y_{0}\in\hat{C}_{1-\alpha}(X_{0})\right\}\ \ge\ 1-\alpha.
		\]
	\end{theorem}
	
	For multivariate covariates, we estimate these conditional quantities with a metric distributional random forest, avoiding the repeated local fits and restrictive single index assumption in \citet{zhou2025conformal}.
	
	\subsection{Metric distributional random forests}
	
	We adapt the distributional random forest of \citet{cevid2022distributional} from Euclidean responses to responses in $(\Omega,d)$ by choosing a kernel on this space.
	
	\subsubsection{Conditional mean embeddings for metric-space responses}
	
	Let $k: \Omega \times \Omega \to \mathbb{R}$ be a bounded, continuous, characteristic kernel. Its reproducing kernel Hilbert space (RKHS) $\mathbb{H}$ has feature map $\varphi: \Omega \to \mathbb{H}$ satisfying $k(y, y') = \langle \varphi(y), \varphi(y') \rangle_{\mathbb{H}}$. For $x \in \mathcal X$, the conditional law $P_{Y \mid X=x}$ has mean embedding \citep{muandet2017kernel, park2020measure}
	\begin{equation*}
		\mu(x) = \mathbb{E}[\varphi(Y) \mid X = x] = \int_{\Omega} \varphi(y) \dl \mathbb{P}(y \mid X=x).
	\end{equation*}
	Because $k$ is characteristic, $\mu(x)$ uniquely determines the conditional law. We estimate this embedding using the forest weights described below.
	
	The choice of kernel depends on the response space. For a general metric $d$, the Gaussian form $\exp\{-\lambda d^2(y,y')\}$ need not be positive definite. This issue arises, for example, with the affine-invariant Riemannian distance between symmetric positive-definite matrices \citep{jayasumana2013kernel}. Appendix~S.2 gives valid kernels for the response spaces considered in Sections~\ref{sec:simulation_studies} and \ref{sec:real_data_analysis}.
	
	\subsubsection{Estimation via metric distributional random forests}
	Maximum mean discrepancy (MMD) compares distributions through the distance between their kernel mean embeddings \citep{gretton2006kernel}.
	Following \citet{cevid2022distributional}, we use MMD to select tree splits. For a node $P$, a covariate coordinate $j$, and a threshold $c$, let
	$
	I_L=\{i:X_i\in P,\ X_i^{(j)}\le c\}
	$
	and
	$
	I_R=\{i:X_i\in P,\ X_i^{(j)}>c\}
	$
	denote the index sets of the two daughter nodes, with $n_L=|I_L|$, $n_R=|I_R|$, and $n_P=n_L+n_R$. The empirical splitting criterion is
	\begin{align*}
		&\frac{n_L n_R}{n_P^2} \cdot \widehat{\mathrm{MMD}}_k^2\left(\{Y_i:i\in I_L\},\{Y_i:i\in I_R\}\right)
		=\frac{n_L n_R}{n_P^2} \cdot \left\| \frac{1}{n_L} \sum_{i \in I_L} \varphi(Y_i) - \frac{1}{n_R} \sum_{i \in I_R} \varphi(Y_i) \right\|_{\mathbb{H}}^2\\
		=&\frac{n_L n_R}{n_P^2} \cdot\left\{\frac{1}{n_L^2} \sum_{i, \ell \in I_L} k\left(Y_i, Y_\ell\right)+\frac{1}{n_R^2} \sum_{i, \ell \in I_R} k\left(Y_i, Y_\ell\right)-\frac{2}{n_L n_R} \sum_{i \in I_L} \sum_{\ell \in I_R} k\left(Y_i, Y_\ell\right)\right\}.
	\end{align*}
	We choose the candidate split $(j,c)$ that maximizes this criterion and repeat the process until the prescribed minimum node size is reached. With a characteristic kernel, MMD responds to differences in the entire response distribution, including changes in spread or shape. CART splitting rules \citep{breiman2001random, qiu2024random} reduce variance or Fr\'echet variance and primarily target conditional centers. Exact evaluation of the MMD criterion is computationally demanding because it involves pairwise kernel evaluations;  the approximations in Appendix~S.2 allow each coordinate to be scanned in linear time after sorting and feature construction.
	
	For simplicity, take $|\mathcal{D}_{\mathrm{tra}}| = |\mathcal{D}_{\mathrm{cal}}| = n$ and $\mathcal{I}_{\mathrm{tra}}=\{1, \ldots, n\}$. Fit a forest of $B$ trees to $\mathcal{D}_{\mathrm{tra}}$. Tree $b$ uses a subsample $\mathcal{D}_{\mathrm{tra}}^b$ of size $s_n$ and randomization $\xi_b \sim \Xi$. Let $\mathcal{L}(x; \mathcal{D}_{\mathrm{tra}}^b, \xi_b)$, abbreviated as $\mathcal{L}_b(x)$, denote the indexed collection of covariates of the observations used for averaging in the terminal leaf containing $x$. Here $|\mathcal{L}_b(x)|$ denotes the corresponding sample size. The forest weights are
	\begin{equation*}
		w_x(X_i) = \frac{1}{B} \sum_{b=1}^B \frac{\mathbb{I}\{X_i\in\mathcal{L}_b(x)\}}{|\mathcal{L}_b(x)|}, \quad i=1, \ldots, n,
	\end{equation*}
	where $\sum_{i=1}^n w_x(X_i) = 1$. The corresponding estimate of the conditional mean embedding is
	\begin{equation}\label{forest embedding}
		\begin{aligned}
			\hat{\mu}_n(x)=\mu\big(\hat{P}_{Y | X=x}\big)
			&=\sum_{i=1}^n w_x(X_i)\varphi(Y_i)\\
			&=\frac{1}{B}\sum_{b=1}^B\left(\frac{1}{|\mathcal{L}_b(x)|}
			\sum_{i:X_i\in\mathcal{L}_b(x)}\varphi(Y_i)\right).
		\end{aligned}
	\end{equation}
	Thus, $\hat{\mu}_n(x)$ is the kernel mean embedding of the weighted empirical measure $\hat{P}_{Y \mid X=x} = \sum_{i=1}^n w_x(X_i) \delta_{Y_i}$, where $\delta_{Y_i}$ denotes the point mass at $Y_i$. For any integrable measurable function $f$ on $\Omega$, we estimate $\mathbb{E}\{f(Y)\mid X=x\}$ by $\sum_{i=1}^n w_x(X_i) f(Y_i)$.
	
	\begin{algorithm}[t!]
		\caption{ELR-based split conformal prediction for object-valued responses}
		\label{alg:split_conformal}
		\begin{algorithmic}[1]
			\Require Observations $\{(X_i,Y_i)\}_{i=1}^{2n}$, miscoverage level $\alpha\in(0,1)$, locality level $\beta\in(0,1]$, and a new covariate $X_{0}$
			\Ensure Prediction set $\hat{C}_{1-\alpha}(X_{0})$ with nominal coverage $1-\alpha$
			\State Randomly partition $\{1,\ldots,2n\}$ into training indices $ \mathcal{I}_{\mathrm{tra}}$ and calibration indices $ \mathcal{I}_{\mathrm{cal}}$; set $\mathcal{D}_{\mathrm{tra}}=\{(X_i,Y_i):i\in \mathcal{I}_{\mathrm{tra}}\}$ and $\mathcal{D}_{\mathrm{cal}}=\{(X_i,Y_i):i\in \mathcal{I}_{\mathrm{cal}}\}$.
			\State Fit a metric distributional random forest to the training set $\mathcal{D}_{\mathrm{tra}}$.
			\State Use the forest weights to compute $\hat{F}_{x}(\omega, r)$, $\hat{R}_{\beta}(\omega \mid x)$, and $\hat S(z \mid x)$.
			\State For each $i\in\mathcal{I}_{\mathrm{cal}}$, compute the nonconformity score $\hat{S}_i = \hat{S}\{\hat{R}_{\beta}(Y_{i} \mid X_{i}) \mid X_{i}\}$.
			\State Let $\hat{Q}_{1-\alpha}$ be the $(1-\alpha)(1 + 1 / |\mathcal{D}_{\mathrm{cal}}|)$ empirical quantile of $\{\hat{S}_i:i\in\mathcal{I}_{\mathrm{cal}}\}$.
			\State Return
			$$
			\hat{C}_{1-\alpha}(X_{0}) \leftarrow
			\Big\{\omega\in\Omega: \hat{S}\big(\hat{R}_{\beta}(\omega \mid X_{0}) \mid X_{0}\big)\le \hat{Q}_{1-\alpha}\Big\}.
			$$
		\end{algorithmic}
	\end{algorithm}
	
	Applying these weights to metric-ball indicators gives the CDP estimator
	$$
	\hat F_{x}(\omega, r)=\sum_{i=1}^n w_{x}\left(X_i\right) \cdot \mathbb{I}\left\{d(Y_i, \omega) \leq r\right\}.
	$$
	Integrating the corresponding distance quantiles gives the CELR estimator
	$$
	\hat{R}_{\beta}(\omega \mid x) = \frac{1}{\beta} \int_0^\beta \hat{Q}_x(\omega, \tau) \, \dl\tau=\frac{1}{\beta} \int_0^\beta \hat{F}^{-1}_x(\omega, \tau) \, \dl\tau.
	$$
	We estimate the conditional distribution function of CELR by
	$$
	\hat{S}(z \mid x)=\sum_{i=1}^n w_x\left(X_i\right) \cdot \mathbb{I}\left\{\hat{R}_\beta\left(Y_i \mid X_i\right) \leq z\right\}.
	$$
	All three estimators use weights from the same forest, so no additional forest fits are required. Algorithm~\ref{alg:split_conformal} summarizes the split conformal procedure.
	For any candidate $\omega\in\Omega$, membership in $\hat C_{1-\alpha}(X_0)$ is determined by
	$
	\hat{S}\big(\hat{R}_{\beta}(\omega \mid X_{0}) \mid X_{0}\big)\le \hat{Q}_{1-\alpha}
	$.
	In numerical work, this rule is evaluated on a finite candidate set suited to the response space.
	
	\section{Theoretical guarantees}\label{sec:theory}
	
	We derive uniform convergence rates for the forest estimators and use them to establish asymptotic conditional validity.
	Following \citet{wager2018estimation, athey2019generalized, cevid2022distributional}, we consider the infinite forest obtained as the number of trees $B$ tends to infinity. It averages over all subsamples of $\mathcal{D}_{\mathrm{tra}}$ of size $s_n$ and over the tree randomization $\xi$ for each subsample. Its weights are
	$$
	w_x(X_i) = \binom{n}{s_n}^{-1} \sum_k \E_{\xi \sim \Xi} \left[ \frac{\mathbb{I}\left\{X_i\in\mathcal{L}(x; \mathcal{D}_{\mathrm{tra}}^k, \xi)\right\}}{|\mathcal{L}(x; \mathcal{D}_{\mathrm{tra}}^k, \xi)|} \right],
	$$
	where $k$ indexes the $\binom{n}{s_n}$ subsamples of $\mathcal{D}_{\mathrm{tra}}$, and the expectation is over $\xi$. We use the same notation $w_x(\cdot)$ for the weights of finite and infinite forests when no ambiguity arises. Writing $Z_i = (X_i, Y_i)$ for $i = 1, \ldots, n$, a generic estimator based on the infinite forest takes the form
	\begin{equation}\label{infinity forest}
		U_{n,s_n}(x) = \binom{n}{s_n}^{-1} \sum_{1 \le i_1 < \ldots < i_{s_n} \le n} \E_{\xi \sim \Xi}\left[ T\left(x; Z_{i_1}, \ldots, Z_{i_{s_n}}, \xi\right) \right],
	\end{equation}
	where $T(x; Z_1, \ldots, Z_{s_n}, \xi)$ is the prediction from a tree built on $\{Z_1, \ldots, Z_{s_n}\}$ with randomization $\xi$. Equation~\eqref{infinity forest} is a $U$-statistic of order $s_n$, which grows with $n$. 
	Working with the infinite forest is standard in \citep{wager2018estimation, athey2019generalized} because it separates statistical error from the Monte Carlo error due to a finite number of trees.
	The latter can be made negligible in practice by choosing $B$ sufficiently large.

	\subsection{Uniform consistency of the forest estimator}
	
	We first prove uniform convergence of the MDRF estimator in RKHS norm. The proof extends the pointwise analysis of \citet{cevid2022distributional} using bounds on leaf diameters and a discretization of the covariate space. We assume the following conditions on the forest construction \citep{wager2018estimation, athey2019generalized}.
	
	\begin{assumption}[Forest regularity]\label{ass:forest regularity}
		The forest satisfies the following properties:
		\begin{itemize}
			\item[(P1)] (Subsampling) Each tree is built from a subsample of size $s_n$ drawn without replacement from the $n$ training observations, where $s_n\to\infty$ and $s_n/n\to0$.
			\item[(P2)] (Honesty) Each tree subsample is randomly divided into equally sized splitting and estimation sets. Tree construction may use the splitting observations and estimation covariates, but not estimation responses, which are used only for leaf averages.
			\item[(P3)] (Random split) 
			At each internal node, the probability of selecting the $j$th covariate for splitting 
			is at least $\pi/p$ for some $\pi>0$ and every $j=1,\ldots,p$.
			
			\item[(P4)] ($\phi$-regularity) 
			Split thresholds are selected from observed covariate values, and each child node receives at least a fraction $0<\phi\le 0.2$ of the estimation observations in its parent node. Trees are grown until every terminal leaf contains between $\kappa$ and $2\kappa-1$ estimation observations for some fixed $\kappa \in \mathbb{N}$.
			
			\item[(P5)] (Symmetry) The distribution of the randomized tree output is invariant under permutations of the training observations.
		\end{itemize}
	\end{assumption}
	
	We also require the following condition on the marginal covariate density, as in Assumption~(A3) of \citet{zhou2025conformal}.
	
	\begin{assumption}[Bounded covariate density]
		\label{ass:bounded_covariate_density}
		The covariates $X \in \mathcal{X}$ have a density that is bounded above and bounded away from zero.
	\end{assumption}

	\begin{theorem}[Uniform consistency of MDRF] \label{thm:dmrf_uniform}
		Let $\mathbb{H}$ be the RKHS associated with a bounded, continuous, and characteristic kernel $k$. Suppose Assumptions~\ref{ass:forest regularity} and~\ref{ass:bounded_covariate_density} hold, and let the subsample size satisfy $s_n \asymp n^\rho$ for some $0<\rho<1$. If the target map
		$$x \mapsto \mu(x) = \E[\varphi(Y)\mid X=x]$$ 
		is Lipschitz continuous, then
		$$
		\sup_{x \in \mathcal{X}} \left\| \hat{\mu}_n(x) - \mu(x) \right\|_{\mathbb{H}} = O_p\left( n^{-\gamma} \sqrt{\log n} \right),
		$$
		where $\gamma = \frac{1}{2} \min\left(1-\rho, \frac{\log(1-\phi)}{\log(\phi)} \frac{\pi}{p} \cdot \rho\right)$.
	\end{theorem}
	
	The rate reflects a bias--variance tradeoff between subsampling variability, represented by $1-\rho$, and the cell-shrinkage rate under $\phi$-regular random splitting, represented by $\frac{\log(1-\phi)}{\log(\phi)}\frac{\pi}{p}\cdot\rho$. The factor $\sqrt{\log n}$ is the cost of extending pointwise convergence to uniform convergence over the covariate space.
	Theorem~\ref{thm:dmrf_uniform} gives uniform control of the conditional mean embedding and hence of the forest-induced conditional law in the MMD metric. This result justifies reusing the forest weights to estimate the distributional quantities required by the ELR-based conformal procedure.
	It does not, however, directly establish consistency of the conditional distance profiles, which are metric-ball probabilities defined by nonsmooth indicator functions.
	
	\subsection{Asymptotic conditional validity for conformal prediction}
	
	To analyze the CRS, we first establish uniform consistency of the CDP estimator over covariates, candidate centers, and radii. This requires regularity in the covariates and control of the complexity of metric balls in the response space.
	
	\begin{assumption}[Lipschitz continuity]\label{ass:Lipschitz continuity}
		The CDP $F_x(\omega,r)$ is Lipschitz continuous in $x$: there exists a constant $L_F>0$ such that, for all $x,x'\in\mathcal X$, $\omega\in\Omega$, and $r\ge0$,
		$$
		|F_x(\omega,r)-F_{x'}(\omega,r)| \le L_F \|x-x'\|_2 .
		$$
	\end{assumption}
	
	\begin{assumption}[Complexity]\label{ass:complexity}
		Let
		$
		\mathcal{G}=\{y \mapsto \mathbb{I}\{d(y, \omega) \leq r\}: \omega \in \Omega, r \geq 0\}.
		$
		There exist constants $A_0<\infty$ and $0<V<2$ such that, for every probability measure $\nu$ on $\Omega$,
		$$
		\log N\left(u, \mathcal{G}, L_2(\nu)\right) \leq A_0 u^{-V}, \quad 0<u \leq 1 .
		$$
	\end{assumption}
	
	Assumption~\ref{ass:Lipschitz continuity} bounds the variation of conditional ball probabilities with $x$; compare Assumption~(A7) of \cite{zhou2025conformal}.
	Assumption~\ref{ass:complexity} controls the uniform entropy of the class of ball indicators. It holds for VC-type ball classes, including those in finite metric spaces, Euclidean spaces and their subsets, finite-dimensional Hilbert spaces, spheres with geodesic distance, symmetric positive-definite matrices with the log-Euclidean metric, and network spaces with the Frobenius metric.

	\begin{theorem}[Uniform convergence of CDP]\label{thm:cdp_uniform}
		Suppose Assumptions~\ref{ass:forest regularity}--\ref{ass:complexity} hold and the subsample size satisfies $s_n \asymp n^\rho$ for some $0<\rho<1$. Let $\hat{F}_{x}(\omega, r)$ denote the MDRF estimator of CDP. Then
		\begin{equation*}
			\sup_{x \in \mathcal{X}, \omega \in \Omega, r \geq 0} |\hat{F}_x(\omega, r) - F_x(\omega, r)| = O_p\left( n^{-\gamma} \sqrt{\log n} \right),
		\end{equation*}
		where $\gamma = \frac{1}{2} \min\left(1-\rho, \frac{\log(1-\phi)}{\log(\phi)} \frac{\pi}{p} \cdot \rho\right)$.
	\end{theorem}
	
	Theorem~\ref{thm:cdp_uniform} gives the same rate as Theorem~\ref{thm:dmrf_uniform}, now uniformly over $(x,\omega,r)$. Uniformity over $x$ follows from the discretization argument used for Theorem~\ref{thm:dmrf_uniform}, whereas control over $(\omega,r)$ relies on a maximal inequality for growing-order $U$-processes. Enlarging the index set does not change the polynomial rate; the factor $\sqrt{\log n}$ accounts for uniformity.
	
	\begin{remark}
		For local linear CDP estimation, \citet[Theorem~2]{zhou2025conformal} obtain stochastic terms $\{(|\log h_n|+\log n)/(nh_n)\}^{1/2}$ and $(nh_n^{1+V/2})^{-1/2}$
		under polynomial uniform covering numbers and polynomial uniform entropy, respectively, where $h_n$ is the bandwidth and $0<V<2$ is the entropy exponent. The second bound contains an additional factor $h_n^{-V/4}$ beyond the usual $(nh_n)^{-1/2}$ scale. In their analysis, the entropy bound is combined with localization to a shrinking covariate neighborhood, so response complexity enters the bandwidth balance and leads to a slower bound as $V$ increases.
		Our forest analysis controls response complexity through normalized tree averages. Each tree prediction is a convex combination of ball indicators with weights that do not depend on $(\omega,r)$. The resulting kernels are bounded by one and inherit the uniform entropy bound with constants independent of $s_n$. Since $V<2$, the entropy integral is finite, giving a uniform stochastic bound of order $O_p(\sqrt{s_n\log n/n})$; see Supplementary Lemmas~S.4 and~S.3. Thus response complexity affects the constant in this bound, while subsampling and the bias from shrinking leaf regions determine the convergence exponent. These bounds do not establish a general rate advantage for forests.
	\end{remark}
	
	To transfer the CDP bound to CELR, we control the tails of distances from a fixed anchor point. Assumption~\ref{ass:anchor_orlicz} allows unbounded response spaces and holds, for example, for Euclidean responses with bounded conditional location and sub-Weibull noise.
	
	\begin{assumption}[Uniform anchor Orlicz integrability]
		\label{ass:anchor_orlicz}
		There exist an anchor point $\omega_0\in\Omega$ and constants $K_0<\infty$
		and $\nu>0$ such that
		$$
		\sup_{x\in \mathcal X}
		\left\|d(Y,\omega_0)\right\|_{\psi_\nu,x}
		\le K_0,
		$$
		where the conditional Orlicz norm is defined by
		$$
		\left\|Z\right\|_{\psi_\nu,x}
		:=
		\inf\left\{
		K>0:
		\E\left[
		\exp\left\{\left(\frac{|Z|}{K}\right)^\nu\right\}
		\mid X=x
		\right]
		\le 2
		\right\}.
		$$
	\end{assumption}
	
	\begin{theorem}[Uniform consistency of CELR]
		\label{thm:elr_uniform}
		Suppose Assumptions~\ref{ass:forest regularity}--\ref{ass:anchor_orlicz} hold and the subsample size satisfies $s_n \asymp n^\rho$ for some $0<\rho<1$. For a fixed $\beta\in(0,1]$, let $\hat{R}_{\beta}(\omega \mid x)$ denote the MDRF estimator of CELR. Then
		\begin{equation*}
			\sup_{x \in \mathcal{X}} \sup_{\omega \in \Omega} |\hat{R}_{\beta}(\omega \mid x) - R_{\beta}(\omega \mid x)|= O_p\left( n^{-\gamma}(\log n)^{1/2+1/\nu} \right),
		\end{equation*}
		where $\gamma = \frac{1}{2} \min\left(1-\rho, \frac{\log(1-\phi)}{\log(\phi)} \frac{\pi}{p} \cdot \rho\right)$.
	\end{theorem}
	
	Theorem~\ref{thm:elr_uniform} establishes uniform consistency of the CELR estimator over covariates and candidate centers. Relative to Theorem~\ref{thm:cdp_uniform}, the polynomial exponent is unchanged, but the additional factor $(\log n)^{1/\nu}$ appears. This factor arises when uniform control of the distance profiles is transferred to their integrated quantiles without assuming a bounded response space. Lighter conditional tails, corresponding to larger $\nu$ in Assumption~\ref{ass:anchor_orlicz}, reduce the logarithmic penalty, while the forest estimation error continues to determine the polynomial rate.
	This result is essential because the conformal score depends on the estimated CELR and its conditional distribution rather than directly on the CDP. Uniform CELR consistency ensures that replacing $R_\beta(\omega\mid x)$ with $\widehat R_\beta(\omega\mid x)$ does not asymptotically distort the ranking of candidate responses.
	
	Uniform consistency of CELR alone is nevertheless insufficient for conditional validity. Because the conformal cutoff depends on the conditional distribution of CELR, small estimation errors must produce only small changes in score ranks. The following continuity condition rules out nonnegligible shifts of probability mass across score thresholds, also stated in Assumption (A1) of \cite{zhou2025conformal}.
	
	\begin{assumption}[Score continuity]
		\label{ass:score_continuity}
		Let $S(z \mid x) = \mathbb{P}\{R_\beta(Y \mid X) \leq z \mid X=x\}$ denote the conditional distribution function of CELR. Assume the following conditions:
		(1) $S(z \mid x)$ is Lipschitz continuous in $z$ uniformly over $x\in\mathcal X$.
		(2) $S(z \mid x)$ is Lipschitz continuous in $x$ uniformly over $z\ge0$.
	\end{assumption}
	
	\begin{theorem}[Asymptotic conditional validity]
		\label{thm:asym_validity}
		Let $\hat{C}_{1-\alpha}(X_{0})$ be the prediction set constructed by Algorithm~\ref{alg:split_conformal}. Suppose Assumptions~\ref{ass:forest regularity}--\ref{ass:score_continuity} hold and the subsample size satisfies $s_n \asymp n^\rho$ for some $0 < \rho < 1$. Then
		\begin{equation*}
			\mathbb{P}(Y_{0} \in \hat{C}_{1-\alpha}(X_{0}) \mid X_{0}) = 1 - \alpha + o_p(1).
		\end{equation*}
	\end{theorem}
	
	Theorem~\ref{thm:asym_validity} complements the marginal coverage guarantee in Theorem~\ref{thm:marginal}. Exchangeability gives finite-sample marginal validity for any score fitted on the training sample. Under the stated regularity conditions, uniform estimation of the score components also yields coverage approaching $1-\alpha$ conditional on $X_0$.
	
	\section{Simulation studies}\label{sec:simulation_studies}
	
	We assess the finite-sample performance of the proposed ELR-based conformal
	method in simulations designed to evaluate conditional coverage across the covariate domain, prediction-set size, and applicability to non-Euclidean responses. In each experiment, the data
	are randomly divided into equally sized training and calibration samples, and split-conformal
	prediction sets are constructed at the nominal coverage level $1-\alpha=0.9$.
	Throughout this section, ELR denotes the proposed procedure estimated by the MDRF, whereas CPS denotes the conditional profile score procedure of \citet{zhou2025conformal} implemented with local linear smoothing. We consider seven settings in total: Settings~1--2, 4, and 6--7 are presented below, while Settings~3 and~5 are reported in Supplementary Section~S.4. Results are averaged over $100$ Monte Carlo replications; Appendix~S.3 describes the hyperparameter choices.
	
	\subsection{Univariate covariates}\label{sec:sim1d}
	
	We begin with the one-dimensional settings of \citet{zhou2025conformal}, taking $X \sim \mathrm{Unif}([-1,1])$, and compare ELR with CPS. Both scalar and sphere-valued responses are considered.
	
	\noindent{\textbf{Setting 1} (Nonlinear regression with heteroscedastic variability).} Given a covariate $X_i \in [-1,1]$, the response $Y_i \in \mathbb{R}$ is generated as
	$$
	Y_i = f(X_i) + \sigma(X_i)\varepsilon_i,
	\quad
	f(x)=(x-1)^2(x+1),
	\quad
	\sigma(x)=0.1\mathbb I\{x\ge0\}+0.5\mathbb I\{x<0\},
	$$
	where $\varepsilon_i \sim \mathcal{N}(0,1)$.
	Hence, the conditional distribution is more dispersed for $x<0$ than for $x\ge0$.

	\noindent{\textbf{Setting 2} (Nonlinear regression with a bimodal pattern).}
	The response $Y_i \in \mathbb{R}$ is generated from the two-component mixture
	$$
	Y_i \sim \tfrac{1}{2}\, \mathcal{N}(f(X_i)+g(X_i), 0.1^2)+\tfrac{1}{2}\, \mathcal{N}(f(X_i)-0.2 g(X_i), 0.1^2),
	$$
	where $f(x)=(x-1)^2(x+1)$ and $g(x)=2\sqrt{x}\,\mathbb{I}\{x \geq 0\}$. The two components coincide for $x\le0$; for $x>0$, their means separate continuously, yielding a bimodal distribution once the separation is sufficiently large.
	
	\begin{figure}[ht]
		\centering
		\includegraphics[width=0.8\linewidth]{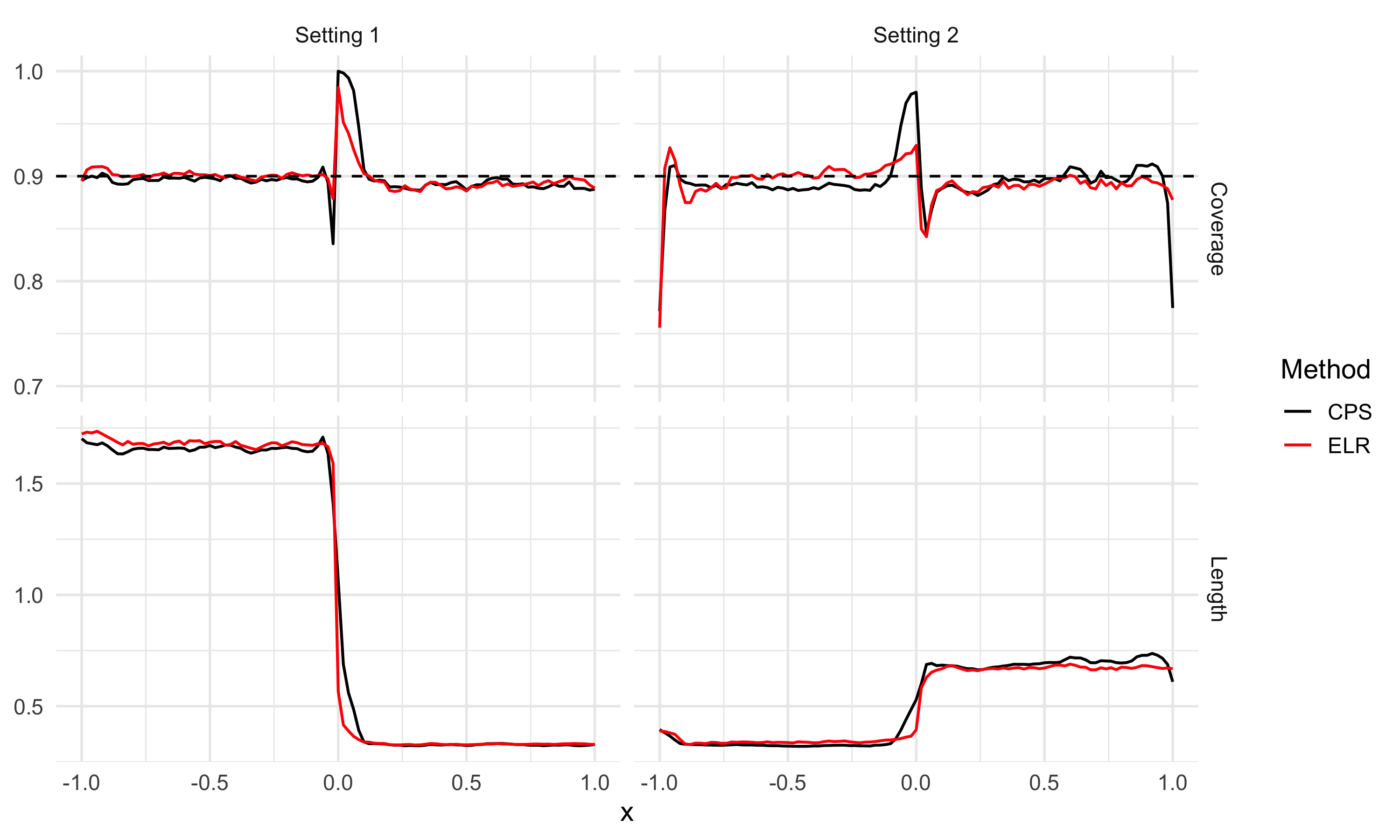}
		\caption{Average conditional coverage (top) and prediction-set length (bottom) under Settings~1--2 with sample size $n=2000$.
			The left and right columns correspond to Settings~1 and~2, respectively.
			Red and black curves denote ELR and CPS, respectively.
			The dashed horizontal line marks the nominal coverage level $0.9$.}
		\label{fig:setting1-2}
	\end{figure}
	
	Figure~\ref{fig:setting1-2} shows empirical coverage close to $0.9$ over much of the covariate range for both methods. In Setting~1, both methods produce longer prediction sets for $x<0$ and shorter sets for $x>0$, reflecting the difference in noise levels. ELR has a sharper decrease in length across $x=0$ and smaller coverage deviations near the discontinuity. 
	In Setting~2, prediction-set length increases near $x=0$ and remains approximately constant for larger positive $x$. ELR shows less overcoverage just to the left of zero, although both methods have undercoverage immediately to its right. Both also show undercoverage at the left endpoint, while the departure at the right endpoint is more pronounced for CPS.
	
	Setting~3, presented in Supplementary Section~S.4.1 and Supplementary Figure~S.1, provides an additional univariate experiment with sphere-valued responses. This experiment shows that both ELR and CPS remain close to the nominal coverage level. Together with Settings~1--2, these results demonstrate that the proposed method remains competitive for univariate covariates with both Euclidean and non-Euclidean responses.

	\subsection{Multivariate covariates}\label{sec:simmulti}
	
	We next consider multivariate covariates
	$
	X=(X^{(1)},\ldots,X^{(p)})^\top \sim \mathrm{Unif}([-1,1]^p).
	$
	These settings assess whether our method can adapt to higher-dimensional covariate structure without explicit parametric assumptions. 
	
	\noindent{\textbf{Setting 4} (Single-index model).}
	For comparison with the multivariate single-index approach of \citet{zhou2025conformal}, let
	$
	\theta=(0.2,0.8,0,\ldots,0)^\top.
	$
	We generate scalar responses from the heteroscedastic single-index model
	$$
	Y_i=f(\theta^\top X_i)+\sigma(\theta^\top X_i)\varepsilon_i,
	\qquad
	\sigma(u)=0.1\mathbb I\{u>0\}+0.5\mathbb I\{u\le0\},
	$$
	where $f(u)=(u-1)^2(u+1)$ and $\varepsilon_i\sim \mathcal{N}(0,1)$.
	
\begin{figure}[!htbp]
		\centering
		\includegraphics[width=0.9\linewidth]{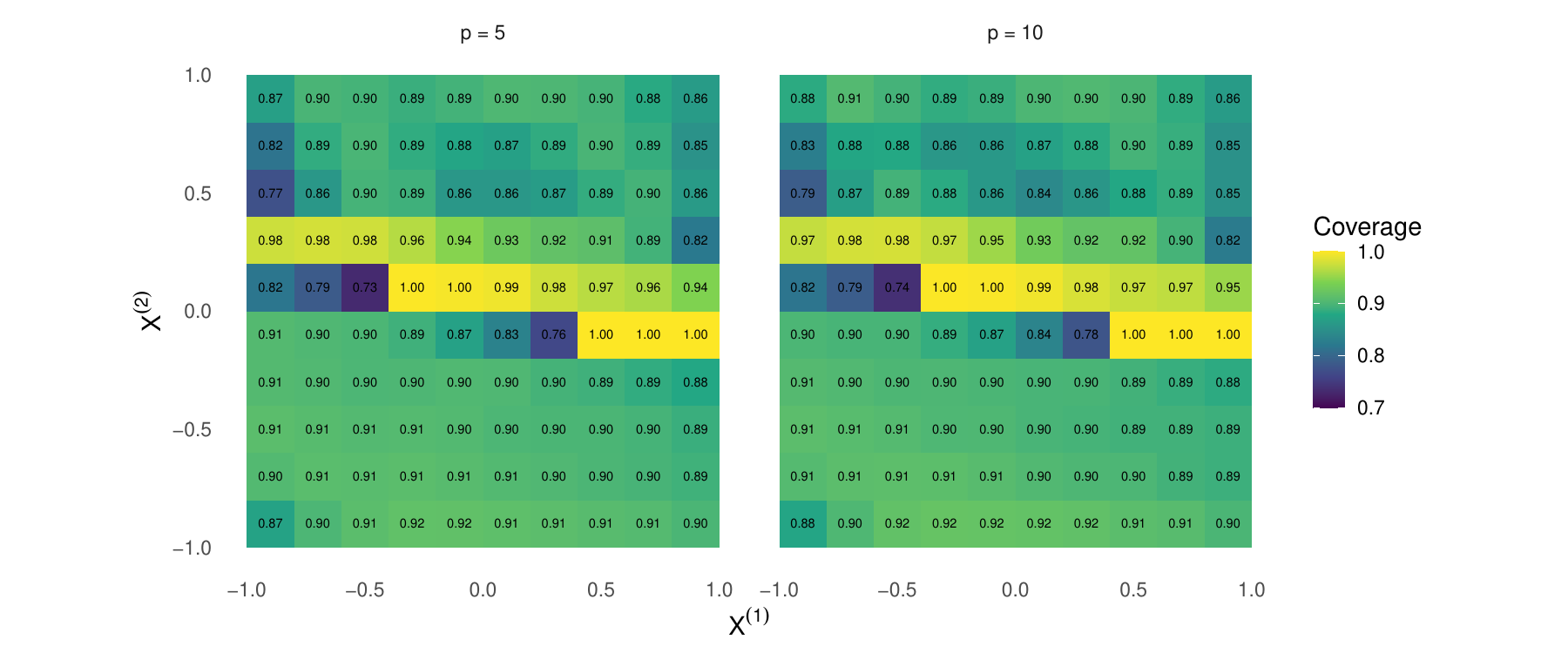}
		\includegraphics[width=0.9\linewidth]{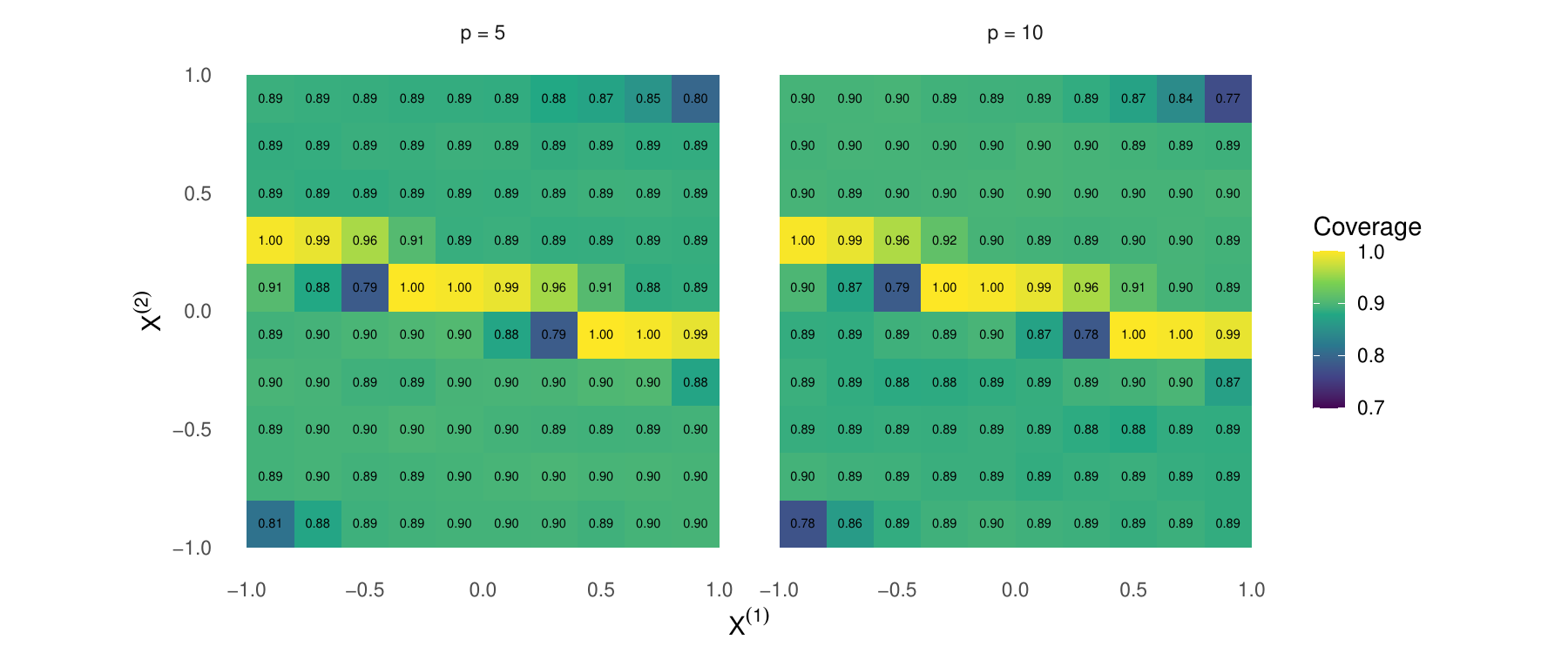}
		\caption{Average conditional coverage for ELR (top) and CPS (bottom) under the single-index model in Setting~4 with $n=2000$. The left and right columns correspond to $p=5$ and $p=10$, respectively. Each panel shows coverage at $100$ test points. The nominal coverage level is $0.9$.}
		\label{fig:setting4}
	\end{figure}

	Figure~\ref{fig:setting4} compares the empirical conditional coverage surfaces of ELR and CPS over a $10\times10$ grid with
	$x^{(1)},x^{(2)}\in\{-0.9,-0.7,\ldots,0.9\}$ and the remaining coordinates fixed at zero. Both methods have coverage close to the nominal level, with local deviations near the change in noise level. CPS, which exploits the correctly specified single-index structure, achieves smaller absolute conditional coverage errors for both $p=5$ and $p=10$. Forest-based ELR offers a nonparametric alternative without imposing this structural assumption, albeit with somewhat larger coverage errors in this experiment. The similar ELR coverage patterns for $p=5$ and $p=10$ suggest little sensitivity to the additional nuisance variables, demonstrating that forest weights are adaptive to the informative coordinates. Setting~5, presented in Supplementary Section~S.4.2 and Supplementary Figure~S.2, provides a complementary multivariate experiment with vector-valued mixture responses. The experiment considers two-component, three-component, and covariate-dependent mixtures; empirical coverage remains close to the target level across these heterogeneous conditional structures. We do not implement CPS in Settings~5--7, as these settings do not satisfy the single-index assumption used in its multivariate extension. 
	
	\noindent{\textbf{Setting 6} (Sphere-valued responses).}
	We next consider sphere-valued responses. Responses
	$Y_i\in\mathbb S^2$ are generated from a von Mises--Fisher distribution,
	$$
	Y_i\mid X_i=x_i \sim \operatorname{vMF}_3(\mu(x_i),\kappa_{\mathrm{vMF}}),
	\qquad \kappa_{\mathrm{vMF}}=60,
	$$
	where
	$$
	\begin{aligned}
		\mu(x)=\left(\sin(\pi g(x)/2),\cos(\pi g(x)/2),0\right)^\top, \,
		g(x)=0.1x^{(1)}+0.2x^{(2)}+0.2(x^{(3)})^2+0.3x^{(4)}x^{(5)}.
	\end{aligned}
	$$
	
\begin{figure}[!htbp]
		\centering
		\includegraphics[width=0.4\linewidth]{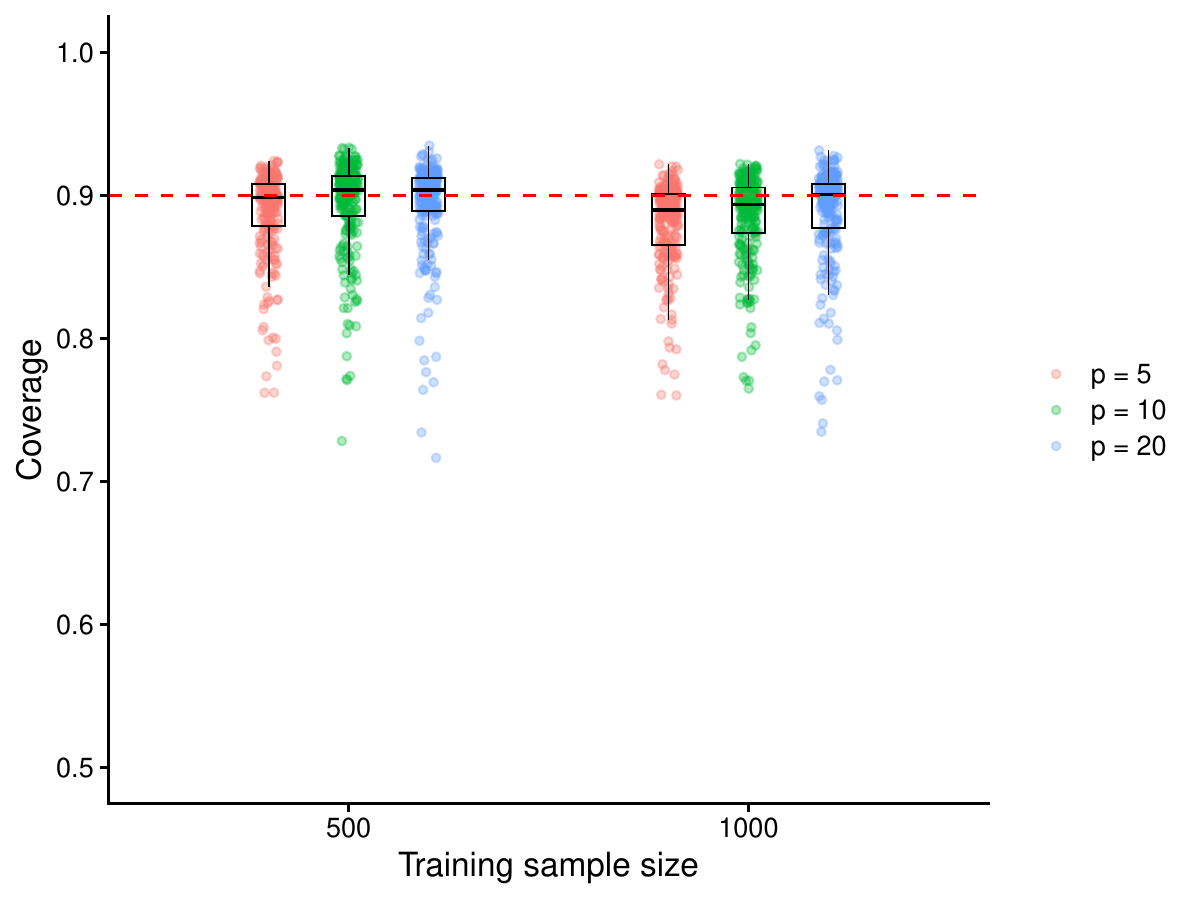}
		\includegraphics[width=0.4\linewidth]{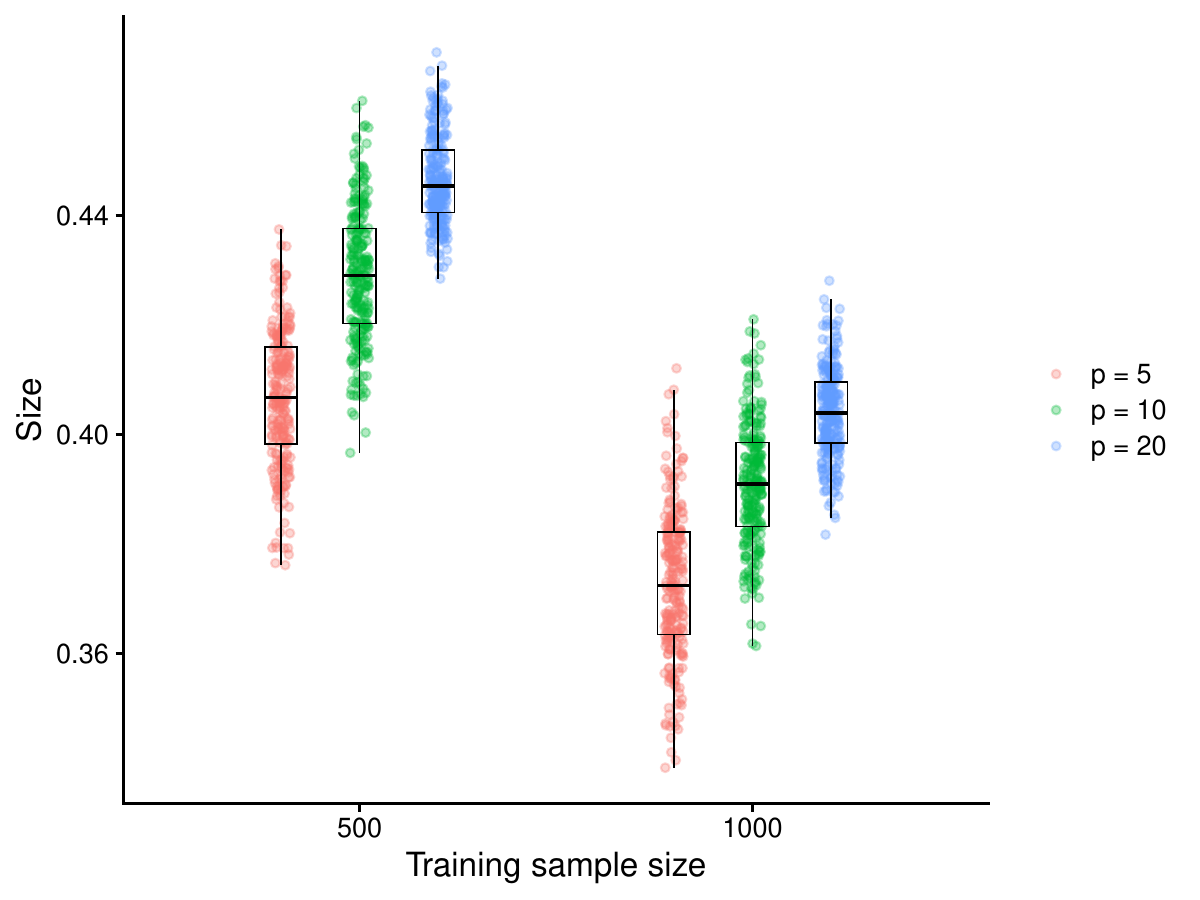}
		\caption{Average conditional coverage (left) and prediction-set size (right) for the sphere-valued model in Setting~6, with $p=5,10,20$ and $n=500,1000$. Each boxplot summarizes the results at $243$ test points. The dashed red line marks the nominal coverage level $0.9$.}
		\label{fig:sphere_multi}
	\end{figure}

	We consider $p=5,10,20$. The test points form a $3^5$ grid over the first
	five coordinates, each taking values in $\{-0.6,0,0.6\}$, with the remaining
	coordinates fixed at zero.
	Figure~\ref{fig:sphere_multi} reports empirical coverage and prediction-set
	size for $n=500$ and $n=1000$. Coverage remains near the nominal level for all three values of $p$. Prediction-set size decreases as the training sample grows, consistent with more accurate estimation of the conditional law.
	Increasing $p$ has only a moderate effect: additional covariates yield slightly larger sets but do not visibly impair calibration. This finding suggests that the forest-based weighting scheme remains adaptive when only a subset of the covariates drives the conditional response distribution.
	
	\noindent{\textbf{Setting 7} (Distribution-valued responses).}
	Finally, we consider distribution-valued responses equipped with the Wasserstein distance. Define
	\begin{gather*}
		\mu(x)
		=
		0.1x^{(1)}
		+
		0.2\sin(\pi x^{(2)})
		+
		0.3x^{(3)}x^{(4)}+\xi_1,\quad \xi_1\sim \mathcal{N}(0,0.5^2),\\
		\sigma(x)
		=
		0.2+0.3|x^{(5)}|+0.1\xi_2,
		\qquad \xi_2\sim{\rm Unif}(0,1).
	\end{gather*}
	The distribution-valued response is
	$$
	Y_i = \mathcal{N}\left(\mu(X_i),\sigma^2(X_i)\right).
	$$
	
\begin{figure}[!htbp]
		\centering
		\includegraphics[width=0.45\linewidth]{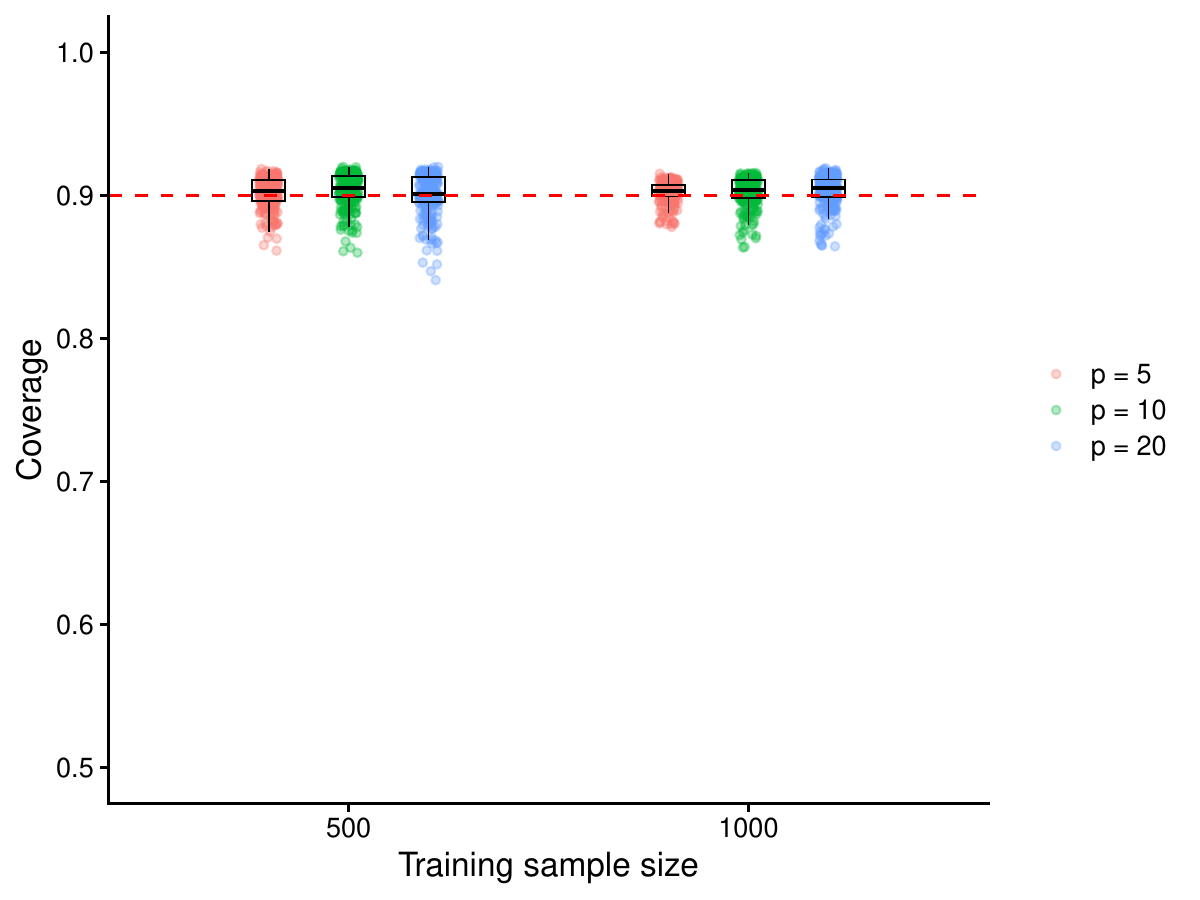}
		\caption{Average conditional coverage for the distribution-valued model in Setting~7, with $p=5,10,20$ and $n=500,1000$. Each boxplot summarizes the results at $243$ test points. The dashed red line marks the nominal coverage level $0.9$.}
		\label{fig:distribution_multi}
	\end{figure}

	We use the same test-point grid as in Setting~6.
	Figure~\ref{fig:distribution_multi} shows that empirical coverage remains stable across the different covariate dimensions and sample sizes. 
	Together with Setting~6, this result demonstrates that the proposed method remains well calibrated for non-Euclidean responses.
	
		\subsection{Computation time comparison}
		\label{sec:runtime}
		
		We compare the computation times of ELR and CPS \citep{zhou2025conformal} in Settings~1, 3, and~4, representing heteroscedastic scalar responses, sphere-valued responses, and a single-index model with multivariate covariates, respectively.
		For each configuration, we run five Monte Carlo replications with preselected tuning parameters held fixed. Within each replication, the methods use the same simulated data, training--calibration split, test covariate points, and candidate-response grid.
		All computations are performed on the same Linux server with Intel Xeon Gold 5218R processors at 2.10~GHz, with both methods restricted to a single CPU thread.		
		We record elapsed time for data generation, model fitting, score evaluation, prediction-set construction, and empirical coverage and size evaluation. 
		
		Table~\ref{tab:runtime} shows that ELR is faster in all three configurations. ELR fits the MDRF once and reuses its weights to estimate CDPs, CELRs, and the conditional distribution functions that define the scores. CPS instead repeatedly applies local smoothing to construct calibration scores and prediction sets, with additional comparisons between conditional distance profiles. The largest timing difference occurs in the multivariate-covariate setting (Setting~4), where CPS also estimates a single-index direction by numerical optimization. MDRF fitting can be further accelerated by parallelizing across trees, although this option is not used in the benchmark.
		
		\begin{table}[!htbp]
			\centering
			\caption{Mean elapsed times (seconds) for ELR and CPS over five Monte Carlo replications. Speedup is the ratio of mean times, \(\overline{T}_{\mathrm{CPS}}/\overline{T}_{\mathrm{ELR}}\).}
			\label{tab:runtime}
			\begin{tabular}{ccccccc}
				\toprule
				Setting & Response space & \(n\) & \(p\) & ELR (s) & CPS (s) & Speedup \\
				\midrule
				1 & \(\mathbb R\)      & 2000 & 1  & 62.2  & 187.7    & \(3.02\) \\
				3 & \(\mathbb S^2\)    & 400  & 1  & 276.8 & 2,515.5  & \(9.09\) \\
				4 & \(\mathbb R\)      & 2000 & 10 & 73.3  & 22,208.7 & \(303.14\) \\
				\bottomrule
			\end{tabular}
		\end{table}

		\section{Real data applications}\label{sec:real_data_analysis}
		
		\subsection{New York City taxi data}
		
		We apply the proposed ELR-based conformal method to data derived from New York City yellow taxi trips recorded during 2024 (\url{https://www.nyc.gov/site/tlc/about/tlc-trip-record-data.page}). This application combines multivariate predictors with symmetric positive-definite matrix (SPD) responses. The trip-level records, released by the New York City Taxi and Limousine Commission, include pickup and drop-off times and locations, trip distance, passenger count, and payment information. We restrict the analysis to Manhattan and partition the borough into 10 subregions. Each response is constructed from a $10\times 10$ network matrix summarizing traffic flows among these subregions.
		Because nighttime traffic is sparse and exhibits distinct patterns, we retain trips between 7{:}00 and 22{:}00 each day and group them into consecutive bins of 1000 trips. For each bin, we form a flow matrix whose $(i,j)$ entry is the number of trips from subregion $i$ to subregion $j$. We scale the matrix by its largest entry, symmetrize it, and then apply the matrix exponential transformation. This yields an SPD-valued response $Y$ representing the taxi-flow network for that bin. Distances between responses are measured using the log-Cholesky metric \citep{lin2019riemannian}.
		
		The 10 predictors are the time of day, defined as the midpoint of each bin; the within-bin averages of trip distance, passenger count, and tip amount; an indicator distinguishing weekdays from weekends and holidays; and five daily weather summaries from Weather Underground for New York City (\url{https://www.wunderground.com/history/monthly/us/ny/new-york-city/KLGA/date}).
        The weather variables are average temperature, average humidity, average wind speed, average barometric pressure, and total precipitation. The taxi analysis in \citet{zhou2025conformal} uses time of day alone; here we use all ten predictors jointly. We randomly divide the observations into training, calibration, and test sets in the ratio $4{:}4{:}2$. Continuous predictors are standardized, while the binary indicator is left unchanged.
		
\begin{figure}[t!]
			\centering
			\includegraphics[width=0.4\linewidth]{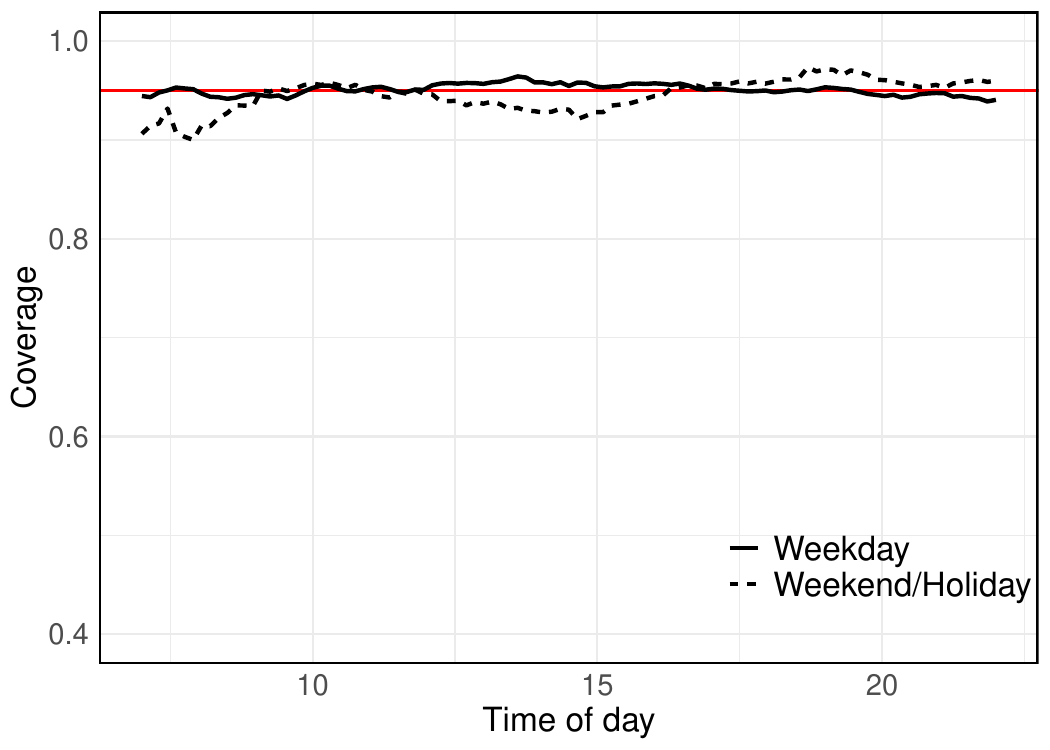}
			\includegraphics[width=0.4\linewidth]{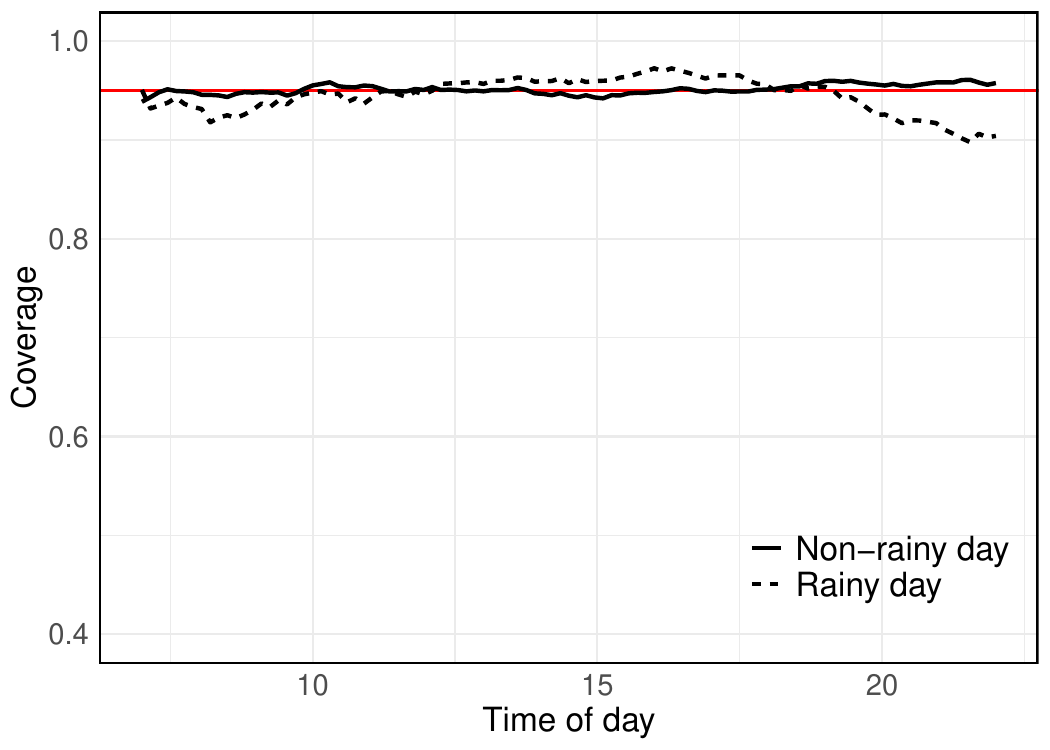}
			\caption{Local empirical coverage as a function of time of day in the New York City taxi application. Left: weekday versus weekend/holiday bins. Right: rainy versus non-rainy conditions. The solid red horizontal line indicates the nominal coverage level $0.95$.}
			\label{fig:taxi_cov_time}
		\end{figure}

		We first evaluate coverage on the test set at the nominal level $1-\alpha=0.95$. The overall empirical coverage is $0.9502$, nearly identical to the target. To examine calibration over the daily traffic cycle, Figure~\ref{fig:taxi_cov_time} plots local empirical coverage against time of day, separately for weekday and weekend/holiday bins and for rainy and non-rainy conditions. Throughout the main daytime period, the curves remain centered near the nominal level, with the larger deviations occurring primarily in sparse or more variable regimes. The method therefore exhibits stable empirical calibration both overall and across substantively different traffic conditions.
		
		To broaden the assessment beyond time of day, we consider one-dimensional diagnostics for representative transportation and weather variables, as well as two-dimensional binned diagnostics for selected covariate pairs. Complete results are reported in Supplementary Section~S.5.1.3 and Supplementary Figures~S.4--S.5. Coverage remains close to the target in most regimes, with the more noticeable departures concentrated in sparse cells or under conditions of high intrinsic variability. These diagnostics support the conclusion from Figure~\ref{fig:taxi_cov_time} that calibration remains stable across diverse traffic and weather conditions, rather than only in aggregate.

\begin{figure}[!htbp]
			\centering
			\includegraphics[width=0.76\linewidth]{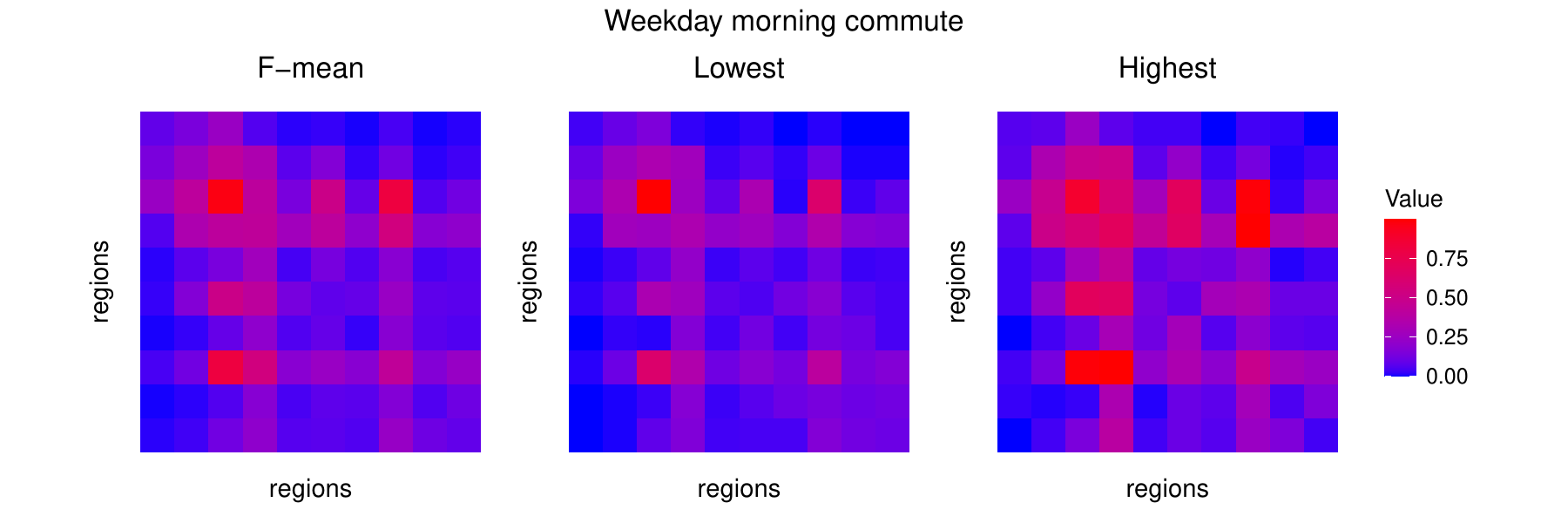}\\
			\vspace{0.2cm} 
			\includegraphics[width=0.76\linewidth]{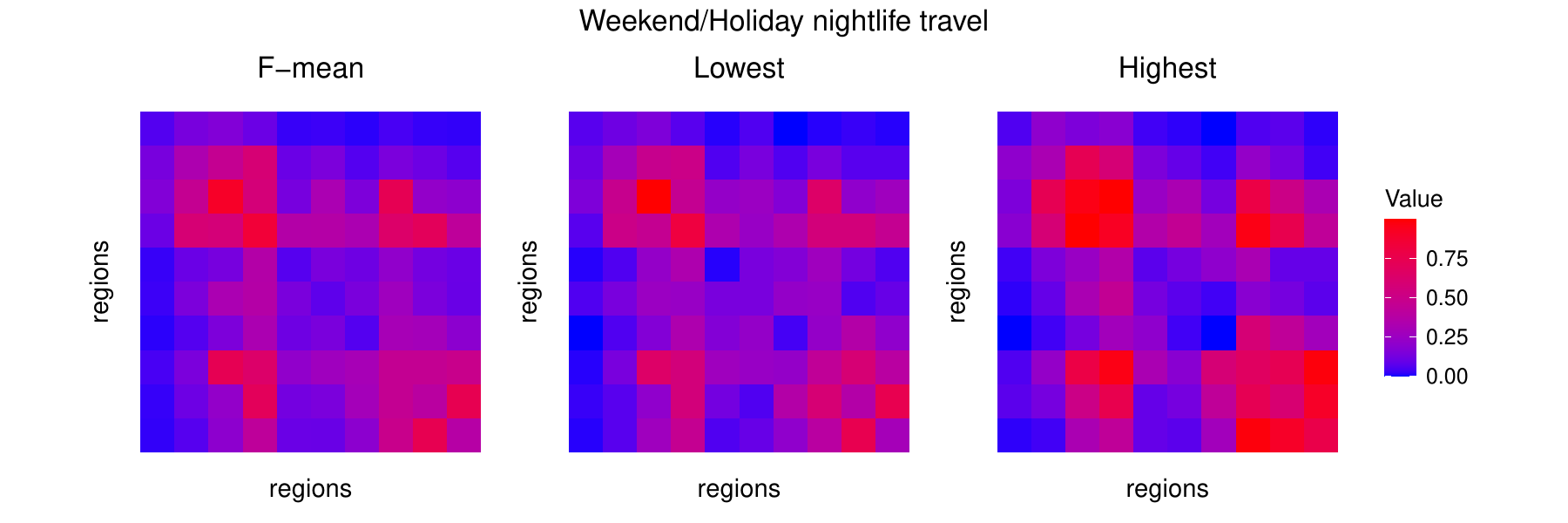}
			\caption{Representative local network patterns in the New York City taxi application. For each traffic regime, we display the local Fr\'echet mean and the observations with the smallest and largest CRS within the corresponding local training sample. Heatmaps display the matrix logarithms of the corresponding SPD matrices.}
	    \label{fig:taxi_representative}
		\end{figure}

        We next use representative local network patterns to examine the interpretation of the fitted CRS. Figure~\ref{fig:taxi_representative} considers two scenarios: weekday morning commuting and weekend/holiday nightlife travel. For each scenario, we display the local Fr\'echet mean and the observations with the smallest and largest CRS within the corresponding local training sample. The local Fr\'echet mean summarizes the selected responses, while the minimum- and maximum-CRS matrices illustrate relatively typical and atypical observations, respectively. Weekday morning networks exhibit more concentrated and recurrent flow patterns, whereas weekend/holiday nightlife networks are more diffuse and less regular. The CRS ordering thus complements the local mean by illustrating variation in conditional typicality within each traffic regime.
		
		Overall, the New York City taxi application illustrates the proposed method in a setting with genuinely multivariate inputs and non-Euclidean outputs. It provides stable empirical coverage for SPD-valued traffic networks together with interpretable local summaries of heterogeneous transportation patterns.

		\subsection{U.S. energy data}
		
		We next analyze a state-level panel of U.S.\ energy consumption (\url{https://www.eia.gov/state/seds/seds-data-complete.php?sid=US}), comprising annual observations for the 50 states and the District of Columbia from 1970 through 2023. For each state-year, the response records the shares of total energy consumption in three broad categories,
		$
		Y=(U,V,W)^\top,
		$
		where $U$ is the share of traditional energy (coal and petroleum), $V$ is the share of natural gas, and $W$ is the share of non-fossil energy, including hydroelectric, wind, solar, geothermal, biomass, and nuclear sources. Since $U+V+W=1$, the response lies on the probability simplex. We map it to the positive orthant of the unit sphere using the square-root transformation
		$
		\widetilde Y=(\sqrt U,\sqrt V,\sqrt W)^\top,
		$
		and construct prediction sets on the sphere using the geodesic metric. This representation preserves the compositional structure of the response while placing the conformal procedure in a natural non-Euclidean space. Our objective is to quantify predictive uncertainty in a changing and heterogeneous energy system, rather than to provide point predictions alone. 
		
		The $10$ predictors are calendar year, population, per-capita electricity consumption, heating degree days, cooling degree days, the nominal prices of coal, petroleum, natural gas, and electricity, and a coastal indicator. These variables represent the long-run energy transition, climate-related demand, incentives for fuel substitution, and persistent geographic heterogeneity. The energy analysis in \citet{zhou2025conformal} uses calendar year alone; our additional covariates allow comparisons among states within the same year. Table~S.4 in Supplementary Section~S.5.2 provides detailed definitions of the response components and covariates. We reserve five benchmark years (1970, 1985, 1999, 2012, 2023) spanning the sample period for testing and divide the remaining observations equally between training and calibration. Continuous predictors are standardized before model fitting, while the coastal indicator remains binary.  

\begin{figure}[!htbp]
			\centering
			\includegraphics[width=0.9\linewidth]{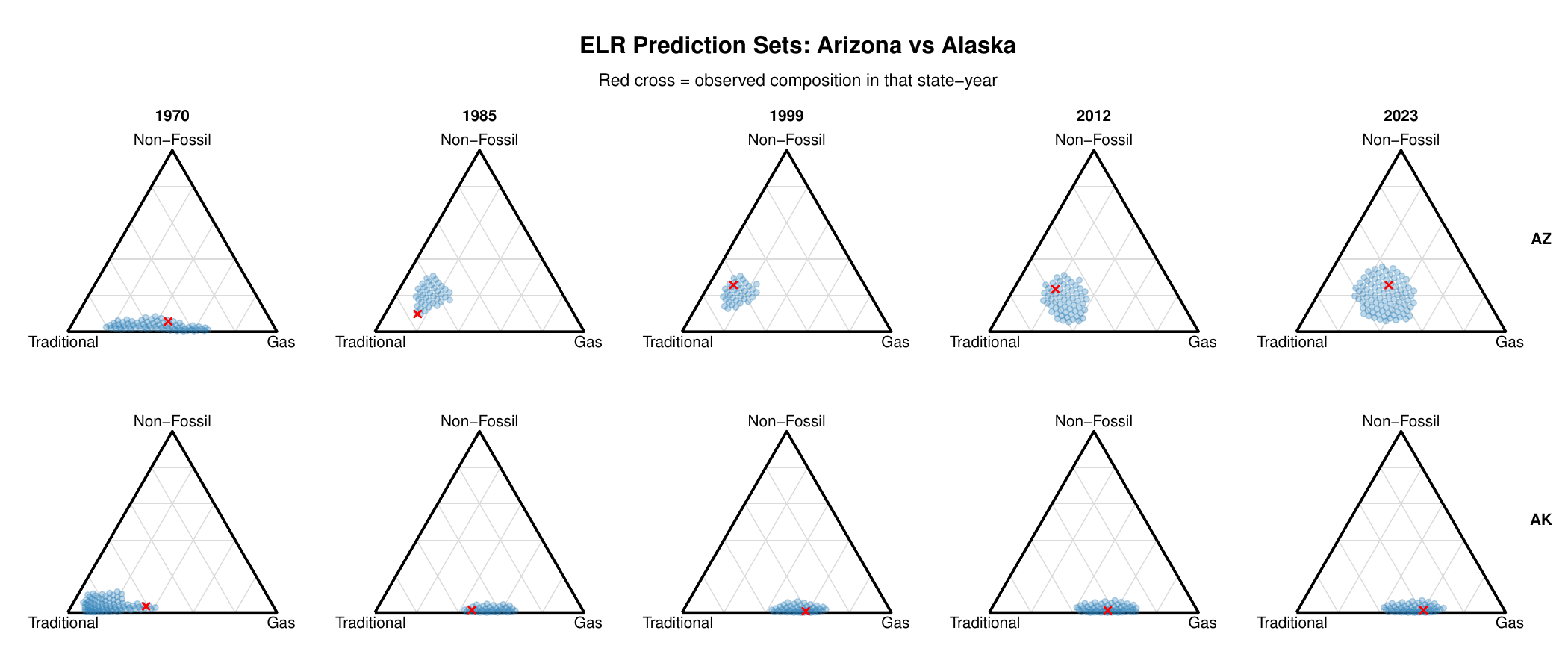}
			\caption{Arizona versus Alaska in the U.S. energy data. Blue regions denote ELR-based conformal prediction sets with nominal coverage $0.9$ across selected years; red crosses mark the observed energy compositions.
            }
			\label{fig:az_vs_ak}
		\end{figure}

        We construct prediction sets at the nominal coverage level $1-\alpha=0.9$ and display them on the ternary simplex. In each panel, the shaded region denotes the ELR-based conformal prediction set, and the red cross marks the observed energy composition. Rather than imposing a fixed-width uncertainty band, the method produces adaptive regions whose location, size, and shape vary with the covariates.
	    Figure~\ref{fig:az_vs_ak} compares Arizona and Alaska, two states chosen for their pronounced climatic contrast. Arizona's prediction sets move progressively away from the traditional-energy edge and toward the non-fossil corner, particularly after the mid-1980s, a pattern consistent with persistent cooling demand and abundant solar resources. Alaska's sets remain concentrated near the traditional-energy side throughout the sample period, although the natural-gas share gradually increases within this fossil-dominated mix. This pattern is consistent with sustained heating demand and improvements in natural-gas transportation and delivery infrastructure.
		
		Two complementary comparisons appear in Supplementary Section~S.5.2.2. Supplementary Figure~S.6 contrasts New York and Wyoming, selected to represent low and high per-capita electricity consumption, while Supplementary Figure~S.7 summarizes broader geographic patterns by averaging responses and covariates within coastal and inland groups. These results show that the prediction sets respond both to state-level structural differences and to the common temporal transition evident after geographic aggregation.
		
		Taken together, these results illustrate three practical features of the method. First, the observed compositions generally fall within or near their corresponding prediction sets, supporting sensible uncertainty quantification for compositional responses. Second, the sets shift over time toward larger shares of non-fossil energy, reflecting the long-run transition of the U.S. energy system. Third, they vary systematically across geographic, climatic, and structural contrasts, indicating that the ELR-based conformal procedure responds to meaningful covariate heterogeneity rather than reproducing a global average pattern.

		\section{Discussion}\label{sec:discussion}
		
		CELR provides a way to construct prediction sets from local concentration in a metric space. Transforming CELR by its conditional distribution function gives the score CRS, whose values are comparable across covariate values. CRS ranks candidates by local concentration rather than distance from a point predictor, so the sets can reflect heteroscedasticity and multimodality without requiring a density or vector space structure. All score components are estimated using the weights from a single metric distributional random forest. Split conformal calibration guarantees finite-sample marginal coverage under exchangeability, and we establish asymptotic conditional validity under the stated regularity conditions.
		
		Sample splitting leaves only part of the data for estimating the conditional law. Out-of-bag forest predictions could make fuller use of the sample by scoring each observation with trees fitted without it. A coverage argument would need to account for the dependence among these scores. Conditional local radii may also be useful for anomaly detection, change point testing, and decision problems under distributional ambiguity in metric spaces.

		%
		%
		%


		\bibliographystyle{abbrvnat}
		\bibliography{reference}

	\end{document}